\documentclass[11pt,journal,onecolumn]{IEEEtran}
\usepackage{cite}
\usepackage{color,soul}

\ifCLASSINFOpdf
\usepackage[pdftex]{graphicx}
  \DeclareGraphicsExtensions{.eps}
\else
  \usepackage[dvips]{graphicx}
  \DeclareGraphicsExtensions{.eps}
\fi
\usepackage[cmex10]{amsmath}
\usepackage[tight,footnotesize]{subfigure}

\usepackage{caption}
\usepackage{amsfonts,amssymb}
\usepackage{epsfig}
\usepackage{algorithm}
\usepackage{algpseudocode}
\usepackage{latexsym}

\DeclareMathOperator{\Div}{div}

\DeclareMathOperator{\median}{median}
\newcommand{\vect}[1]{\boldsymbol{#1}}
\newcommand{\ip}[2]{<#1 , #2 >}
\newcommand{\ipG}[2]{<#1, #2>_G}
\newcommand{\cTV}[1]{\int_{\Omega} \vert \nabla #1 \vert}

\begin{document}
%
\title{Enhanced Compressed Sensing Recovery with Level Set Normals}
%
%
%

\author{Virginia Estellers$^\star$, Jean-Philippe Thiran$^\star$, Xavier Bresson$^\circ$
\thanks{V. Estellers and J.P Thiran are with the Signal Processing Laboratory (LTS5), Ecole Polytechnique F\'ed\'erale de Lausanne.}
\thanks{X. Bresson is with the Department of Computer Science, City University of Hong Kong.}}

\maketitle

\begin{abstract}
We propose a compressive sensing algorithm that exploits geometric properties of images to recover images of high quality from few measurements. The image reconstruction is done by iterating the two following steps: 1) estimation of normal vectors of the image level curves and 2) reconstruction of an image fitting the normal vectors, the compressed sensing measurements and the sparsity constraint. The proposed technique can naturally extend to non local operators and graphs to exploit the repetitive nature of textured images in order to recover fine detail structures. In both cases, the problem is reduced to a series of convex minimization problems that can be efficiently solved with a combination of variable splitting and augmented Lagrangian methods, leading to fast and easy-to-code algorithms. Extended experiments show a clear improvement over related state-of-the-art algorithms in the quality of the reconstructed images and the robustness of the proposed method to noise, different kind of images and reduced measurements.
\end{abstract}



%


\section{Formulation of the problem}
\IEEEPARstart{C}{ompressed} sensing (CS) is founded on the principle that, through optimization, the sparsity of a signal can be exploited to recover it from a reduced number of measurements. This simple and yet powerful idea is intriguing because it seems to violate Shannon's sampling theorem. Compressed sensing is in fact the equivalent of Shannon's theorem from the point of view of sparsity: while Shannon states that to recover a band limited signal the sampling rate must be at least twice the maximum frequency present in the signal; CS relates the sparsity of a signal in certain basis with the number of measurements in another basis necessary to recover it from an $\ell_1$ minimization problem.  A few definitions are necessary to understand the formulation of the CS problem.

We say that a signal $u\in \mathbb{R}^n$ is $s$-sparse in the basis or dictionary $\Psi$ if it can be expressed by $s$ non-zero coefficients in that basis, i.e. $\Vert \Psi u\Vert_0 = s$; while $u$ is compressible if most of the energy in $\Psi u$ is contained in its largest $s$ coefficients. Given $\Phi$ and $\Psi$ two orthobasis or dictionaries of $\mathbb{R}^n$, the CS proplem is formulated as the reconstruction of a signal $u \in\mathbb{R}^n$, sparse in basis $\Psi$, from $m<n$ linear measurements $f$ in the sensing basis $\Phi$. Ideally we should measure the $n$ projections of $u$ in basis $\Phi$, that is $\Phi u$, but we only observe a small subset $f= A u$ of size $m < n$. The sampling matrix $A= R \Phi$ results from the combination of the sensing basis $\Phi$ and the matrix $R$ that extracts the corresponding to the measurements in $f$.  Consequently, the system $f = A u$ is undetermined and the sparsity of the signal $u$ must be exploited to ``invert'' the problem and obtain a correct reconstruction. The obvious strategy would be to recover the sparsest $u$ agreeing with the measurements, that is, to solve the following non-convex problem 
\begin{align}
\min_{u} \Vert \Psi u \Vert_0 \hspace{4pt} \textrm{s.t} \hspace{4pt} Au = f . 
\label{eq:l0}
\end{align}
Problem \eqref{eq:l0} is NP-hard due to the $\ell_0$ norm and only approximate solutions can be used in real applications. Relaxing the $\ell_0$ norm to $\ell_1$, problem \eqref{eq:l0} becomes the convex problem
\begin{align}
\min_{u} \Vert \Psi u \Vert_1 \hspace{4pt}  \textrm{s.t} \hspace{4pt}  Au = f .
\label{eq:l1}
\end{align}
One of the key results in CS \cite{Candes2006} proves that \eqref{eq:l1} exactly recovers $s$-sparse signals with an overwhelming probability when the number of measurements $m\geq c \log{n}$ (with small constant $c$). In addition, if the sampling matrix $A$ verifies certain restricted isometry condition, then \eqref{eq:l1} actually recovers the signal associated to the $s$ largest coefficients of $u$ in basis $\Psi$, i.e. exact recovery for $s$-sparse signals and recovery of the s-sparse $\ell_2$ approximation for compressible signals. 

When the measurements are contaminated with noise, the constraint $Au=f$ on the measurements is relaxed. In particular, under Gaussian noise the recovery is given by 
\begin{align}
\min_{u} \Vert \Psi u \Vert_1 \hspace{4pt}  \textrm{s.t} \hspace{4pt}  \Vert Au - f \Vert_2 \leq \sigma
\label{eq:l1noisy},
\end{align}
 where $\sigma$ is related to the noise level. From optimization theory \cite{Boyd2010}, we know that \eqref{eq:l1noisy} is equivalent to
 \begin{align}
\min_{u} \Vert \Psi u \Vert_1 +\frac{\alpha}{2} \Vert Au - f \Vert_2 ^2 
\label{eq:l1noisy_unconstraint}
\end{align}
 in the sense that solving any of the two determines the parameter ($\sigma$,$\alpha$) in the other and both have the same minimizer. 
 
Designing the sparsifying basis depends on the signal at hand. For images a common choice are orthogonal wavelets or the discretized total variation (TV) seminorm. TV assumes that the edges of an image are sparse and it is extensively used in inverse imaging problems as a regularizer. In its continuous formulation TV is a convex functional and its usual discretizations preserve that property. In CS $\Vert \Psi u \Vert_{1}$ is then substituted by the regularizing term $J(u) = \Vert u \Vert_{BV}$ in an abuse of notation.

Without loss of generality in this paper we adopt the lagrangian formulation \eqref{eq:l1noisy_unconstraint}, use random Fourier samples as measurements \footnote{The proposed matrix $A$ satisfy the restricted isometry condition with high probability and is therefore a comon choice in MRI imaging \cite{Rudelson2006}. } and choose total variation as sparsity criterion; but the proposed algorithm could be equally applied to other basis or dictionaries as investigated in \cite{lustig2005,Candes2005,Chakraborty2008}. The CS recovery problem that we consider is then
\begin{align}
\min_{u} J(u) + \frac{\alpha}{2} \Vert Au - f \Vert_2 ^2 .
\label{eq:CSmin}
\end{align}
With this formulation of CS recovery in imaging, we introduce an additional term in \eqref{eq:CSmin} inspired by image denoising techniques \cite{Lysaker2004}. The resulting model exploits the geometry of the image to improve image recovery by aligning the normals associated to the levels sets of the image with the reconstructed signal. Our first contribution is therefore the introduction of a term for CS recovery based on geometric properties intrinsic to images. Our method can be beautifully extended to non local operators in order to recover textured images. In this case we exploit the geometry of the graph defined by the non local operators to recover finer details and structures of the images. This observation is a key contribution of our work because it can be easily adapted to improve existing non local denoising and deblurring methods, not only CS recovery. Finally, it is also important to mention that the proposed CS recovery model is based in the solution of two convex optimization problems and therefore can be efficiently solved with fast and easy-to-code algorithms.

The rest of the paper is organized as follows. After formulation the problem in this section, we present our method in Section \ref{sec:local_method} and explain its relation to similar techniques in Section \ref{sec:relatedapproaches}. Our method is then extended to non local operators in Section \ref{sec:NL_method}. Section \ref{sec:minlocal} presents the associated minimization problems.  Finally, experiments are presented in Section \ref{sec:experiments} and conclusions drawn in Section \ref{sec:conclusions}.

\section{CS with recovered normals} \label{sec:local_method}
The main idea behind our method is that the recovered normals of an image can significantly improve the CS recovery results. This observation raises two main questions: how to recover normals robustly and accuratly from CS measurements and how to introduce the estimated normals in the CS recovery. The answer that we propose is a two-step iterative method. 

In the first step of each iteration, we estimate the normal vectors\footnote{We use boldface for vector fields.} $\vect{n}$ associated to the level set curves of the image by solving a vectorial TV model. Once the normals are estimated, we find an image that fits the measurements, the estimated normals and the sparsity criterion. The process is then iterated and can be   
summarized as
\begin{equation}
\left\{
  \begin{array}{ l}
\vect{n}_{k} = \arg \min_{ \vert \vect{n}\vert \leq 1 } J_{w}(\vect{n}) + \frac{\mu}{2} \Vert \vect{n} - \vect{\hat{n}} \Vert_2 ^2 \\
u_{k} =  \arg \min_{u} J(u) +\gamma \ip{ \Div \vect{n}_{k} }{ u}  + \frac{\alpha}{2} \Vert Au - f \Vert_2 ^2 \label{eq:CSnormal}
\end{array} \right.
\end{equation}
On the following subsections we will detail each of these two steps, which both reduce to convex optimizations that can be efficiently solved. Combining the two stages into one would lead to a non convex model of higher order and the resulting minimization would be slower and suffer from local minima. A two step method is computationally more efficient in the same way than splitting variables in Section \ref{sec:minlocal} helps solving the minimization problems and leads to closed form solutions. The drawback of this two step procedure is the lack of rigorous theory and proof of convergence of the resulting algorithm. However, experimental results show that a single iteration of our method already improves the standard recovery \eqref{eq:CSmin}, while the peak performance is attained after a few iterations.

\subsection{Estimation of level set normals}\label{sec:normals1}
Each iteration, the normals of the image are estimated in two steps. We first obtain a noisy point-wise estimate $\vect{\hat{n}}$ from the previous solution $u_{k-1}$ and we then regularize it to obtain $\vect{n}_k$. 
\\
Given an estimate of the underlying image $u_{k-1}$, the normal vectors associated to its level set curves are defined as
\begin{align}
\vect{\hat{n}} =  \frac{ \nabla u_{k-1} }{ \vert \nabla u_{k-1} \vert} .
\label{eq:normals1}
\end{align}
Denoising of that first estimate of the normals $\vect{\hat{n}}$ is done with a combination of the vectorial ROF model \cite{Blomgren1998} with the constraint $\vert \vect{n}\vert \leq 1$. In particular we define the vector field $\vect{n}_k = ( n_x, n_y)_k$ as the solution of the following variational problem. 
\begin{align}
 \min_{ \vert \vect{n}\vert \leq 1 } J_{w}(n_x,n_y) + \frac{\mu}{2} \Vert n_x - \hat{n}_{x} \Vert_2 ^2 + \frac{\mu}{2} \Vert n_y - \hat{n}_{y} \Vert_2 ^2 \label{eq:normROF}
\end{align}
where $J_w(n_x,n_y)$ is the extension of the weighted TV seminorm to vector fields and $w= g( \vert \nabla u_{k-1} \vert )$ is an edge detector designed to verify $w \approx 0$ near the edges and $w \approx 1$ on flat regions of $u_{k-1}$. 
\\
By weighting the TV seminorm with an edge detector $w = g\left(\vert \nabla u_{k-1}\vert \right)$, we encourage the edges of the regularized solution to coincide with the main edges of the noisy signal $u_{k-1}$. To be robust against false edges, we use the robust edge detector proposed by Black, Sapiro and Marimont in \cite{Black1998}, where an statistical interpretation of the edge-stopping functions of anisotropic diffusion \cite{Perona1990} is given. In this statistical interpretation, edges are considered outliers in the normal distribution of $\vert \nabla u \vert $ associated to noisy piece-wise constant regions and the edge-stopping functions $g\left(\vert \nabla u \vert \right)$ are derived from error norms robust to outliers. The edge detectors therefore have a parameter $\sigma$ that acts as a soft-threshold in the detection of outliers and can be estimated a priori from the values of $\vert \nabla u \vert $ in the image. 
Based on the results of \cite{Black1998}, we define 
\begin{align}
g(x) = \begin{cases} \frac{1}{2}\left( 1 - \frac{x}{\sigma}^2\right)^2  & \vert x \vert \leq \sigma \\ 0 & \vert x \vert > \sigma \end{cases}
\label{eq:edge_detector}
\end{align} 
with $\sigma =1.4826 \median ( \vert  \nabla u - \median ( \vert  \nabla u  \vert ) \vert )$.
\\
The constraint $\vert \vect{n}\vert \leq 1$ in \eqref{eq:normROF} corresponds to a relaxation of the condition $\vert \vect{n}\vert =1$ inherent to the definition of normals. It is numerically necessary in flat regions, where $\nabla u = \vect{0}$ and we cannot numerically normalize the gradient vector.

\subsection{Matching normals and CS measurements}

Once the normal field $\vect{n}_k$ is computed, we find an image that matches this field by including the term $-\ip{\vect{n}_k}{\nabla u}$ in the standard CS recovery model \eqref{eq:CSmin}. This term tries to maximize the alignment of the estimated normals of the signal $\vect{n}_k$ with the normals of the reconstruction $ \frac{\nabla u}{\vert \nabla u \vert }$. The resulting recovery model is
\begin{equation}
u_{k+1} =  \arg \min_{u} J\left( u \right) -\gamma \ip{ \vect{n}_{k+1}}{ \nabla u} + \frac{\alpha}{2} \Vert Au - f \Vert_2 ^2  \hspace{8pt}.
\end{equation}
Taking into account that the divergence $\Div$ is the adjoint operator of the gradient $\nabla$, the previous minimization can be rewritten as 
\begin{equation}
u_{k+1} =  \arg \min_{u} J\left(  u \right) +\gamma \ip{\Div \vect{n}_{k}}{u} + \frac{\alpha}{2} \Vert Au - f \Vert_2 ^2  
\label{eq:min}.
\end{equation}
Our method then exploits the geometry of the image in the recovery process and obtains better regularization properties than standard TV. In particular the proposed model preserves edges like TV, by encouraging the gradients to be sparse with $J(u)$; but is also able to recover smooth regions by aligning the gradients of the reconstruction with the smoothed normals with the term $\ip{ \vect{n}_{k+1}}{ \nabla u}$.

In principle we could also use a smooth estimate of the gradients $ \vect{v} = \nabla u_{k-1} $ instead of $\vect{n}=\frac{ \nabla u_{k-1} }{ \vert \nabla u_{k-1} \vert}$ to align the gradients of the reconstructed signal. However, discontinuities of the image would have a contribution to $\vect{v}$ proportional to their jump and the resulting term $\ip{\vect{v}}{ \nabla u}$ would give different weights to discontinuities of different sizes. From a geometric perspective, if we want to recover the shapes of the image independently of their contrast we need to consider the normal vectors derived from its level sets. By the use of level sets, we treat all shapes equally and the term $\gamma \ip{ \vect{n}_{k+1}}{ \nabla u}$ only accounts for geometric quantities.

\section{Related approaches in CS} \label{sec:relatedapproaches}
The method that we propose is inspired by image denoising and impainting methods \cite{Lysaker2004,Ballester2001} that align an estimate of the normals with the reconstructed image. In the context of image denoising, Lysaker, Osher and Tai in \cite{Lysaker2004} first regularize the unit gradient of the noisy image and then improve reconstruction by fitting this gradient into the regularized vector. The resulting method outperforms the ROF model \cite{Rudin1992} and similar higher order PDE methods \cite{Lysaker2003}. Dong et al. in \cite{Dong2009} improve this model by regularizing the angles instead of the vectors and introducing an edge indicator as an extra weight. In image inpainting, an equivalent two-step method is proposed by Ballester et al. in \cite{Ballester2001}, later improved with the divergence free constraint by Tai in \cite{Tai2007book} and adapted to image decomposition and denoising in \cite{Hahn2011}. In general, processing the normals to improve reconstruction has also been used in shape from shading\cite{Horn86thevariational} and mesh optimization\cite{Ohtake2001789}. However, this information has not been exploited before for CS image recovery.

In the CS field, several methods have been proposed to improve the quality of the $\ell_1$ recovery. For general signals, greedy algorithms\cite{Tropp2007,Needell2009301,Blumensath2009265}  and $\ell_p \ 0<p<1 $ minimizations \cite{Rao738251,Chartrand2007} approximate the solution of the $\ell_0$ problem \eqref{eq:l0} and improve its sparsity; but the resulting minimizations are not convex, the algorithms are slow and suffer from local minima. To improve the sparsity of $\ell_1$ recovery \eqref{eq:CSmin} without increasing its complexity, Cand\`es, Wakin and Boyd \cite{Candes2008a} proposed an iterative process solving a weigthed $\ell_1$ problem at each iteration. The weights are defined inversely proportional to the value of the recovered signal in the previous iterate, approximating the behaviour of the $\ell_0$ norm and promoting sparser signal recovery. The resulting method efficiently solves a convex problem at each iteration, experimentally improves signal recovery and has been adopted for image processing with TV regularization in the edge-guided CS of Guo and Yin\cite{Guo2010}.  Edge-guided CS incorporates information about the magnitude of the gradient in the recovery process and it is therefore realted to our method. However, we propose an additive method more robust to noise and exploit both the magnitude and directional information of the gradients.

CS recovery of images has also been improved modifying the data term $ \Vert Au - f \Vert_2 ^2$ inspired by image denoising techniques. In particular, the Bregman iterations proposed by Osher et al. in \cite{Osher2005} for image denoising and deblurring have been applied to CS in \cite{He2006a}. He et al. in \cite{He2006a} use Bregman iterations to improve CS image recovery for phantom MRI data, but fail in the recovery of real images due to the additional difficulties of reconstructing a signal from partial measurements compared to the original denoising problem. For the particular case of TV regularisation, the first Bregman iteration has a geometric interpretation similar to the second step of our recovery method. However, Bregman iterations do not include a regularization step for the normals and therefore fail for noisy and real MRI signals. 

In the following, we summarize each of these to methods and clarify their relationship with our technique.

\subsection{Edge-guided CS}
Edge-guided CS \cite{Guo2010} improves recovery of MRI images by exploiting edge information with an iterative process that weights TV with an edge detector associated to the image recovered in the previous iteration. The key idea is that edges correspond to locations where $\vert \nabla u \vert$ is large, TV corresponds to the $\ell_1$ norm of the norm of the gradient and therefore an inverse edge detector can be used to re-weight TV and approximate the $\ell_0$ norm in a similar fashion to the re-weighted $\ell_1$ of Cand\`es, Wakin and Boyd \cite{Candes2008a} for general signals. The method starts with the standard CS solution \eqref{eq:CSmin} to obtain a first estimate of the image $u_1$ and its edges. It then defines the weights $w_1 = g(\vert \nabla u_1 \vert ) $ inversely proportional to $\vert \nabla u_1 \vert )$ in order to recover an image with sparser edges at the second iteration by solving the re-weighted TV problem. The process is iterated, leading to the following two step algorithm:  
\begin{align}
\left\{
  \begin{array}{ l}
    u_{k+1} =  \arg\min_{u} J_{w_k}(u) + \frac{\alpha}{2} \Vert Au - f \Vert_2 ^2\\
    w_{k+1} = g( \vert \nabla u_{k+1} \vert )
  \end{array} \right.
\label{eq:edgeCS}
\end{align}
There is no stopping criterion or guarantee of convergence for this iterative process and usually, after a few iterations the reconstruction does not improve or even degrades. In fact, the  multiplicative model of edge-guided CS is very sensitive to false edge detection. In particular, if an edge is detected in a wrong location, the weight associated to it on the next iteration will encourage an edge on this location and CS recovery will degrade with any new iterations. The iterative re-weighting process is designed to improve sparsity of the signal and recovery of piecewise constant functions, but it fails in the recovery of smooth regions in images. Compared to our method, edge-guided CS incorporates only information about the magnitude of $\nabla u$, while we also use its directional information; it does not include a regularization step for the detected edges and it is specially designed for piecewise constant images.

\subsection{Bregman methods}
We also share similarities with Bregman methods, whose original idea was to restore normals and image intensity simultaneously. However, Bregman methods cannot recover normals as accuratly and robustly as our method because they do not regularize the estimated normals. Our improvement is at the price of loosing global convexity.

Bregman iterations substitute the minimization problem \eqref{eq:CSmin} for a sequence of convex optimizations substituting $J(u)$ for its Bregman distance to the previous iterate. In particular, the first Bregman iteration has a geometric interpretation closely related to our method. 
Starting with $u=0,v=0$, the Bregman iterative process can be summarized as
\begin{equation}
\left\{
  \begin{array}{ l}
u_{k+1} =  \arg\min_{u} J(u) + \frac{\alpha}{2} \Vert f + v_k - A u\Vert_2 ^2 \\
v_{k+1} = v_k +  f - Au_{k+1}
  \end{array} \right.
\label{eq:IRM}\end{equation}
While their first iteration corresponds to the standard CS model \eqref{eq:CSmin}, their second iteration implicitly exploits the normals of the image recovered at iteration one to improve the recovery. For simplicity, here we show the connection to our method with the continuous formulation, where $A(\cdot)$ is the continous functional operator of CS and $A^{*}$ its adjoint. 
For the first iteration $u=0,v=0$ and the method solves
\begin{align}
u_{1} =  \arg\min_{u\in\mathbb{R}^n} \cTV{u} + \frac{\alpha}{2} \Vert f - A(u) \Vert_2 ^2, \label{eq:IRM1}
\end{align}
The optimality condition associated to \eqref{eq:IRM1} derived from its the Euler-Lagrange equation is
\begin{align}
\Div \frac{\nabla u_1}{ \vert \nabla u_1 \vert} =  - \alpha A^{*}(u_1) \left( f - A(u_1) \right) \label{eq:optcond}
\end{align}
 where $\vect{n}_1 = \frac{\nabla u_1}{ \vert \nabla u_1 \vert} $ correspond to the normals of $u_1$. At the next iteration we can introduce a term $\ip{\vect{n}_1}{\nabla u}$ aligning the normals of the reconstructed signal with the estimate of the normals from the previous iteration, that is
 \begin{align}
u_{2} =  \arg\min_{u} \cTV{u}  - \ip{\vect{n}_1}{\nabla u} +  \frac{\alpha}{2} \Vert f - A(u) \Vert_2 ^2
\label{eq:IRM2}
\end{align}
Integrating by parts and substituting $\Div \vect{n}_1$ in \eqref{eq:optcond} we  have
\begin{align}
 - \ip{\vect{n}_1}{\nabla u}  = \ip{\Div \vect{n}_1}{u} =  -\ip{ \alpha A^{*} \left( f - A(u_1) \right)}{u}  = \nonumber \\- \alpha \ip{ f - A(u_1)}{A(u)} = - \alpha \ip{ v_1}{A( u)}
 \label{eq:IRMaux}
\end{align}
with $v_1 =  f - A(u_1)$ as defined in \eqref{eq:IRM}. If we susbstitute \eqref{eq:IRMaux} in the minimization \eqref{eq:IRM2} and group together the terms with $A(u)$, we end up with the Bregman update rule 
\begin{align}
u_{2} =  \cTV{u}  +   \frac{\alpha}{2} \Vert f + v_1 - A(u) \Vert_2 ^2 .
\end{align}
For the rest of iterations the geometric interpretation of the update is lost. Compared to Bregman iterations, our method explicitly uses the normals in the recovery model for all iterations, not only the second one, and it is not restricted to TV regularization. Indeed, this geometric interpretation is only possible for the TV term $J(u)$, while our method can be used with any sparsifying basis. We are also more robust to noise thanks to the regularization step and, unlike the Bregman iteration, experimentally improve the reconstruction model \eqref{eq:CSmin} for both phantom and real MRI data. In addition, our method extends to non local operators to exploit graph geometry and recover fine details in textured images.

\section{Extension to non local methods} \label{sec:NL_method}
Total variation regularization is designed to recover images with sharp edges but, as other methods based on local gradients, it is not suited for textured images with fine structures. In this section we extend our method to textured images using both a non local TV regularization and a term aligning the estimated non local normals with the non local gradients of the reconstructed image.
\subsection{Non local operators}
Non local TV is a variational extension of the non local means filter proposed by Buades, Coll and Morel for image denoising\cite{Buades2005}. Non local means exploits the repetition of patterns in natural and textured images to reconstruct sharp edges as well as fine meaningful structures. That principle is the basis of non local regularization methods in imaging, which outperform the classical methods by incorporating global information in the regularization process. In \cite{Gilboa2008} Gilboa and Osher use graph theory to extended the classical TV to a non local functional. In the discrete setting, Zhou and Sch\"olkopf \cite{Zhou2005} and Elmoataz et al. \cite{Elmoataz2008} use graph Laplacians to define similar non local regularization operators. The resulting non local methods have been applied to image denoising \cite{Gilboa2008}, segmentation \cite{gilboa:595,Bresson2008a}, inpainting \cite{Peyre2008}, deconvolution and compressive sensing \cite{Zhang2010}. 

We adopt the discrete formulation of the continuous model presented in \cite{Gilboa2008}. In this non local framework we consider the image domain as a graph $G = (\Omega, E)$; where $\Omega$ is the set of nodes of the graph, one for each pixel in the image, and $E$ is the set of edges connecting the nodes. The edge connecting nodes $i$ and $j$ is weighted with a positive symmetric weighting function $w(i,j)$ that represents the distance between the two nodes in graph terms. Consequently, two pixels $i$ and $j$ spatially far away in the image can be considered neighbours in the graph and interact if $w(i,j)>0$ (we write then $i\sim j$). For that reason, the resulting approach is considered non local. 

Given an image $u$ defined on the graph, the non local gradient $\nabla_G u$ at node $i$ is defined as the vector of all directional derivatives associated to the neighbours of $i$, that is  
\begin{align}
\nabla_G u \ (i,j) = ( u(j) - u(i) )\sqrt{w(i,j)} \hspace{4pt} \forall j\in \Omega .
\end{align}
In the graph, vectors $\vect{d} = d(i,j)$ are therefore functions defined in the domain $\Omega\times\Omega$.
\\
In this setting we define the standard $L_2$ inner product between functions as
\begin{align}
\ipG{u}{v} = \sum_{i\in\Omega} u(i) v(i) .
\end{align}
For vectors, we define a dot product pixel-wise 
\begin{align}
 (\vect{d} \cdot \vect{e})_G (i)= \sum_{j\sim i} d(i,j) e(i,j)
\end{align}
and an inner product on the graph
\begin{align}
 \ipG{\vect{d}}{\vect{e}}= \sum_{i} (\vect{d} \cdot \vect{e})_G (i) = \displaystyle\sum_{i} \displaystyle\sum_{j\sim i} d(i,j) e(i,j) .
\end{align}
In order to have an equivalent to the TV seminorm, we define a norm function on the graph $\vert \cdot \vert_G$. With the previous definitions, the magnitude of a vector at node $i$ is given by
\begin{align}
\vert \vect{d} \vert_G (i) = \sqrt{ (\vect{d} \cdot \vect{d})_G (i) }= \sqrt{\displaystyle\sum_{j\sim i} d(i,j)^2}
\end{align}
The standard TV is then naturally extended to a non local version as the $\ell_1$ norm of the graph norm $\vert \cdot \vert_G$ associated to the non local gradient, that is, 
\begin{align}
TV_{G}\left( u\right) = J_{G}\left( u\right)  = \sum_{i} \vert \nabla_G u \vert_G (i)   = \Vert \ \vert \nabla_G u \vert_G \ \Vert_1 .
\end{align}
With the above inner products, the non local divergence of a vector $\vect{d}$ is defined as the adjoint of the non local gradient, that is
\begin{align}
 \Div_G \vect{d} \ (i) =  \sum_{j\sim i} ( d(i,j) - d(j,i) )\sqrt{w(i,j)} .
\end{align}
\\
With these definitions, if we consider only the immediate pixels as neighbours and fix their weights to $w(i,j)=1$, then the non local TV reduces to the standard TV definition. If we consider more general neighbours by defining a correct weighting function like in \cite{Buades2005}, the non local operators incorporate global information and the standard regularization process is improved. 
The weight function therefore has an important impact in the performance of the non local regularizers. Inspired by \cite{Buades2005,Gilboa2008}, given a reference image $u_0$ we compute weighting function $w_0(i, j)$ measuring the difference of patches around each node as follows
\begin{align}
 w_0 (i,j) = \exp^{-\frac{\Vert \mathcal{P}_0 (i) - \mathcal{P}_0 (j)  \Vert^2}{2h^2}},
\end{align}
where $h$ is a scaling factor and $\mathcal{P}_0 (i)$ is a patch of $u_0$ centered at pixel $i$.
This weighting function is designed to reduce Gaussian noise while preserving the textures of the image. The reference image should be as close as possible to the true image in order to incorporate valid information related to image structures in the non local operators. For that reason, we initialize the weighting function in the non local methods with the standard CS solution \eqref{eq:CSmin} (on the following $u_0$) and iteratively solve the non local model and update the weights with the non local solution. The basic non local CS recovery is then
\begin{equation}
\left\{
  \begin{array}{ l}
\nabla_{G_k} \longleftarrow \textrm{ estimate non local operators from $u_{k-1}$ }\\
u_{k} =  \arg \min_{u} J_{G_k}(u)  + \frac{\alpha}{2} \Vert Au - f \Vert_2 ^2 . 
\end{array}  \right.\label{eq:NLCSmin}
\end{equation}

\subsection{Proposed non local method}
Symmetrizing our local technique, we propose a two step iterative method for CS recovery. In the first step of each iteration, we estimate the non local normals $\vect{n}_G$ associated to the level set curves of the image in the graph. Once the non local normals are estimated, we find an image that fits the non local normals and the CS measurements and iterate the process. 
\\
In the local setting, the normal vectors associated to the level set curves of an image $u$ are defined as $\vect{n} = \frac{\nabla u}{\vert \nabla u \vert}$. We extend that definition to our non local framework and exploit the geometry of the image in the graph to improve CS recovery. In particular, we derive the equivalent non local normals from the non local gradient $\nabla_{G} u$ by normalizing its components node-wise, i.e. all the components associated to node $i$ are normalized by $\vert \nabla_{G} u \vert_G (i)$.

Given an estimate of the non local normals $\vect{n}_G$, we can include a term in the CS reconstruction \eqref{eq:NLCSmin} maximizing the alignment of the reconstructed signal with the normals. The resulting minimization is
\begin{align}
u =  \arg \min_{u}  J_{G}(u) -\gamma \ipG{ \vect{n}_{G}}{\nabla_G u} + \frac{\alpha}{2} \Vert Au - f \Vert_2 ^2 
\end{align}
Exploiting the adjoint relation of the non local divergence and gradient $\ipG{ \vect{n}_{G}}{\nabla_{G}u} = -\ipG{ \Div_{G}\vect{n}_G}{u}$, we have
\begin{align}
u = \arg \min_{u} J_{G}(u)  +\gamma \ipG{\Div_G n}{u} + \frac{\alpha}{2} \Vert Au - f \Vert_2 ^2 .
\end{align}
As before, the process can be iterated and we obtain the following analogue to the previous two step procedure:
\begin{equation}
\left\{
  \begin{array}{ l}
\nabla_{G_k}\longleftarrow \textrm{ estimate non local operators from $u_{k-1}$} \nonumber\\
\Div{G_k}\vect{n}_{G_k} =  \arg \min_{v} J_{G_k}\left( v\right)  + \frac{\mu}{2}\Vert v - \hat{v} \Vert^2  \nonumber\\
u_{k} =  \arg \min_{u} J_{G_k}(u) +\gamma \ipG{\Div{G_k}\vect{n}_{G_k} }{u}  + \frac{\alpha}{2} \Vert Au - f \Vert_2 ^2 \label{eq:NLCSnormal}
\end{array} \right.
\end{equation}
with $ \hat{v} = \left( 1 - g(\vert \nabla_{G} u_{k-1} \vert_G ) \right) \Div_{G} \frac{\nabla_{G} u_{k-1}}{\vert \nabla_{G} u_{k-1} \vert}$.

The third step of our non local method is naturally derived from our local version and the geometric interpretation of the non local operators. However, the regularization step of the non local normals requires careful consideration, as we explain next.

\subsection{Estimation of non local normals}
The non local gradient operator, and consequently the non local normals, do not correspond to the discretization of standard vector fields in a grid. Indeed, $ \nabla_{G} u$ has a different number of components for each pixel and the associated direction to $ \nabla_{G} u (i,j)$ depend on the relative position of the node $i$ and its neighbour $j$. Therefore, we cannot use standard techniques to regularize these vector fields and we prefer to regularize the term $\Div_{G} \vect{n}$ posteriorly used in the recovery algorithm. Compared to the regularization of the non local normals, we loose directional information, but the resulting method is simpler and faster.
\\
Assume that we are given an estimate of the reconstructed signal $u_{k-1}$. We first compute a noisy estimate of the non local normals and their divergence pixel-wise and we then denoise it with standard denoising methods. In particular, we estimate the non local normals as 
\begin{align}
\vect{\hat{n}_G} = \frac{\nabla_{G} u_{k-1}}{\vert \nabla_{G} u_{k-1} \vert}
\end{align}
and compute a rough estimate of the non local divergence as
\begin{align}
\hat{v} = \left( 1 - g(\vert \nabla_{G} u_{k-1} \vert_G ) \right) \Div_{G} \vect{\hat{n}_G}, 
\label{eq:NLdiv1}
\end{align}
where $g(x)$ is a function designed to verify $g\approx 0$ when $x$ is large and $g\approx 1$ when $x$ is small. In fact, $g(\vert \nabla_{G} u_{k-1} \vert_G )$ acts as the equivalent edge detector presented in Section \ref{sec:normals1} and is defined with the same expression \eqref{eq:edge_detector}. As in the local case, we adopt the statistical interpretation of the edge detector $g\left(\vert \nabla_G u \vert_G \right)$ presented in \cite{Black1998},  where the edges are considered as outliers in the normal distribution of $\vert \nabla_G u \vert_G $ associated to homogeneous regions. Since the edge detector $g$ is derived from error norms robust to outliers, weighting our estimate of the normals with the function $ 1 - g(\vert \nabla_{G} u_{k-1} \vert_G )$ in \eqref{eq:NLdiv1} is equivalent to soft-thresholding the non local normals when we suspect that the variations in $u_{k-1}$ are due to noise inside homogeneous regions.
\\
Finally, we regularize $\hat{v}$ to obtain a smoother estimate of the non local divergence, which will be used in the second step of our iterative method. There are two natural approaches for this regularization: we can ignore the non local nature of the divergence and gradient operators and use any local model to regularize $\hat{v}$, for instance the standard ROF\cite{Rudin1992} of equation \eqref{eq:ROF}; or use the non local neighbours to denoise $\hat{v}$ with \eqref{eq:NLROF}, that is, use the natural distance and neighbouring relations inherent to de definition of $\hat{v}$ to denoise it.
\begin{align}
\Div{G_k}\vect{n}_{G_k} =  \arg \min_{v} J\left( v\right)+ \frac{\mu}{2}\Vert v - \hat{v} \Vert^2 \label{eq:ROF}\\
\Div{G_k}\vect{n}_{G_k} =  \arg \min_{v} J_G\left( v\right)  + \frac{\mu}{2}\Vert v - \hat{v} \Vert^2 \label{eq:NLROF}
\end{align}
In our experiments we obtained slightly better results with the first approach.

\section{Minimization problems} \label{sec:minlocal}
In order to solve the minimization problems involved in each step of our method, we make use of recent advances in convex  minimization\cite{Wang2008,Goldstein2009} and apply variable splitting and augmented Lagragians\cite{Glowinski1981} to obtain efficient and easy-to-code algorithms. To simplify notation on this section we remove the subindexes in $u_k$ and $n_k$ indicating the iterations of our two step procedure.

The minimizations associated to each of the local steps involve both a TV and a quadratic term similar to the ROF model\cite{Rudin1992}. Consequently, the resulting algorithms apply a similar strategy to overcome the non-linearity and non-differentiability of TV than the multitude of algorithms proposed for ROF. In the original ROF paper\cite{Rudin1992}, the authors derive the Euler-Lagrange PDE of the model and propose a time marching method to solve it. The resulting method is slow due to the constraint on the time step associated to its stability conditions. In the last years more efficient methods have been proposed for the ROF model due to its extensive use in imaging. A popular class of methods is based on the dual formulations of the ROF model, e.g. Chambolle's projection method \cite{Chambolle2004} or primal-dual approaches\cite{Chan1999,Zhu2008,ChambollePock:2011}. Other options are based on variable-splitting and equality constrained optimization; which is solved by quadratic-penalties \cite{Wang2008}, Bregman iterations\cite{Goldstein2009,Yin2008} or the equivalent augmented Lagrangian method \cite{Wu2010}. In the case of CS,  dual solvers are not usually adopted because they suffer from matrices $A$ that are large-scale and dense. In particular for matrices corresponding to transforms with fast implementations (like the Fourier transform of this paper), splitting methods are a good choice because they can easily exploit fast transforms to compute $Au$ and $A^{T} u$\cite{Wang2008,Goldstein2009}. The algorithms that we propose fall in this last category. We rewrite the different problems as constraint minimizations and use augmented Lagrangians to solve them. The resulting Lagrangians are minimized with respect to each variable independently and the multipliers are then updated in a cyclic way. Since all the minimizations can be analytically solved, the resulting algorithms are extremely fast and easy to implement.

Similarly, the minimization algorithms that we propose for the non local method is closely related to the minimization of the non local ROF model proposed in \cite{Gilboa2008}, which was originally solved with a time consuming time marching algorithm. The non local CS problem has been solved with a combination of forward-backward splitting and Bregman iteration in \cite{Zhang2010}, but for uniformity of the paper we use the same combination of splitting and augmented Lagrangians than in the local case to solve the non local problem \eqref{eq:NLCSnormal}. 

\subsection{Minimizations of local normal-guided CS}

We discretize the image domain $\Omega \subset \mathbb{R}^2$ with a regular grid of size $n = n_x\times n_y$. In $\Omega$ we consider images as scalar functions with $u(i)\in \mathbb{R}$ and their gradients as vector-valued functions with $\nabla u (i) \in \mathbb{R}^2$. \\
We use forward differences to compute the discrete gradients and backward differences for the divergence in order to preserve the adjoint relationship $\Div = - \nabla^{*}$ in the discrete setting. 
\\
The discrete TV semi-norm is then given by
\begin{align}
J(u) = \sum_{i}\vert \nabla u (i) \vert = \sum_{i} \sqrt{ \nabla_x u (i)^2 + \nabla_y u (i)^2 }
\end{align}
where we denote the pixel-wise norm of vectors as $\vert \vect{d} \vert (i) =\sqrt{ d^2_x(i) + \vert d^2_y(i)} $.
Our discretized TV is then the $\ell_1$ norm of the function computing the pixel-wise norm of the gradient, i.e $J\left( u \right) = \Vert \ \vert \nabla u \vert \ \Vert_1$.  
\\
For vector fields $\vect{d} =\left( d_x, d_y\right)$, we discretize the TV seminorm as follows
\begin{align}
J(d_x,d_y) = \sum_{i} \sqrt{ { \vert \nabla d_x (i) \vert }^2 + {\vert \nabla d_y (i) \vert }^2 }.
\end{align}
In that case we observe that it corresponds to the $\ell_1$ norm of the function computing the pixel-wise norm of the vector of combined gradients, i.e $J\left( d_x, d_y \right) = \Vert \ \vert \left( \nabla d_x, \nabla d_y \right) \vert \ \Vert_1$. With that observation it is easy then to extend it to a weighted TV norm as  $J_w( d_x, d_y ) = \Vert \ \vert W \left( \nabla d_x, \nabla d_y \right) \vert \ \Vert_1$, where $W$ is the diagonal matrix of weights.

In the vector notation used in CS, we can efficiently compute the spatial derivatives multiplying the discrete functions arranged as a column vector with the sparse finite difference matrices $\nabla_x u = D_x u $, $\nabla_y u = D_y u$. Similarly, the discretization of the $L_2$ inner product in $\Omega$ corresponds to the usual dot product of vectors, i.e. $\ip{v}{u} = v^{T} u$. 

\subsubsection{Estimate $u$ from CS measurements and normals} \label{sec:min_local1}
To reconstruct the image we need to solve the following convex minimization problem:
\begin{equation}
\min_{u} \Vert \ \vert \nabla u \vert \ \Vert_1 +\gamma v^{T} u + \frac{\alpha}{2} \Vert Au - f \Vert_2 ^2  \hspace{8pt} \textrm{with}  \hspace{8pt}  v = \Div n\label{eq:minbis}.
\end{equation}
We propose an iterative algorithm to solve \eqref{eq:minbis} based on splitting and constraint minimization techniques. The main idea is to split the original problem into sub-optimization problems which are well-known and easy to solve, and combine them together using an augmented Lagrangian. The proposed algorithm is guaranteed to converge thanks to the convexity of \eqref{eq:minbis}. 

Let us consider the following constrained minimization problem, which is equivalent to \eqref{eq:min}:
\begin{equation}
\min_{u,\vect{d}} \ = \ \Vert \ \vert \vect{d} \vert \ \Vert_1 +  v^{T}u + \frac{\alpha}{2} \Vert Au - f \Vert_2 ^2 \ \textrm{ s.t. }\ \vect{d}= \nabla u 
\label{eq:cmin}
\end{equation}
The proposed splitting approach makes the original problem \eqref{eq:min} easier to solve because \eqref{eq:cmin} decouples the $\ell_1$ norm and the gradient operator $\nabla$.

Next, we reformulate this constrained minimization problem as an unconstrained optimization task. This can be done with an augmented Lagrangian approach, which translates the constraints into pairs of Lagrangian multiplier and penalty terms. Let us define the augmented Lagrangian energy associated to \eqref{eq:cmin}:

\begin{align}
\mathcal{L}_1\left( u, \vect{d} ,\vect{\lambda} \right) = \Vert \ \vert \vect{d} \vert \ \Vert_1 +  v^{T}u + \frac{\alpha}{2} \Vert Au - f \Vert_2 ^2 + \lambda^{T}_{x} ( d_x - D_x u) \nonumber \\ + \lambda^{T}_{y} ( d_y - D_y u)  + \frac{r}{2} \Vert d_x - D_x u \Vert_2 ^2 + \frac{r}{2} \Vert d_y - D_y u \Vert_2 ^2 
\label{eq:AL_local1}
\end{align}
where $\vect{\lambda}=\left(\lambda_x,\lambda_y\right)$ are the Lagrange multipliers and $r$ is a positive constant. 
\\
The constraint minimization problem \eqref{eq:cmin} reduces to finding the saddle-point of the augmented Lagrangian energy $\mathcal{L}_1$. The solution to the saddle point problem \eqref{eq:AL_local1} can be approximated iteratively by the following algorithm: initialize the variables and Lagrange multipliers to zero; at each iteration find an approximate minimizer of $\mathcal{L}_1~(u,~\vect{d},~\vect{\lambda}^{k-1})$ with respect to the variables $u, \vect{d}$ and update the Lagrange multipliers with the residuals associated to the constraints; stop the process when $u$ remains fix. As the Lagrangian $\mathcal{L}_1$ is convex with respect to $u, \vect{d}$, we can find a minimizer  by iteratively alternating the minimization with respect to each variable. The resulting method is equivalent to the alternating direction method of multipliers. The iterative method is summarized in Algorithm~\ref{alg:AL_it}
\begin{algorithm}
\caption{Augmented Lagrangian method to solve \eqref{eq:cmin}, estimating $u$ from CS measurements and normal matching}
\label{alg:AL_it}
\begin{algorithmic}[1]
\State Initialize $ u, \vect{d}, \vect{\lambda}$
\State For each iteration $l=1,2\ldots$, find an approximate minimizer of $\mathcal{L}_1$ with respect to variables $(u, \vect{d})$ with fixed Lagrange multipliers $\vect{\lambda}^l$:
\small{
\begin{align}
u^{l} = \arg\min_{ u } \mathcal{L}_1\left(u, \vect{d}^{l-1},\vect{\lambda}^{l-1} \right) & \textrm{ solved in in Fourier domain} \label{eq:min1}\\
\vect{d}^{l}= \arg\min_{ \vect{d} } \mathcal{L}_1\left(u^{l}, \vect{d},\vect{\lambda}^{l-1} \right)& \textrm{ solved by shrinkage}\label{eq:min2}
\end{align}
}
\State Update Lagrange multipliers
\small{
\begin{align}
\lambda^{l}_x =& \lambda^{l-1}_x + r( d^{l}_x - D_x u^{l} )\nonumber\\
\lambda^{l}_y =& \lambda^{l-1}_y + r( d^{l}_y  - D_y u^{l} )\nonumber
\end{align}
}
\State Stop the iterative process when $\frac{ \Vert u^l - u^{l-1} \Vert}{\Vert u^l \Vert} <\epsilon$.
\end{algorithmic}
\end{algorithm}

The next step is to determine the solutions of the two sub-minimization problems~\eqref{eq:min1},\eqref{eq:min2}, which can be computed efficiently.

The sub-minimization problem \eqref{eq:min1} can be written as follows:
\begin{align}
\min_u \ v^{T}u + \frac{\alpha}{2} \Vert Au - f \Vert_2 ^2 +  \frac{r}{2} \Vert d_x + \frac{1}{r}\lambda_x - D_x u \Vert_2 ^2 + \nonumber \\ \frac{r}{2} \Vert d_y + \frac{1}{r}\lambda_y - D_y u  - \Vert_2 ^2 .\label{eq:en1}
\end{align}
We see that it reduces to a quadratic minimization, with positive semi-definite Hessian $ H=\alpha F^{T}R^{T} R F+r ( D^{T}_x D_x+ D^{T}_y D_y)$.
The optimality conditions read 
\begin{equation}
 H u = b \hspace{4pt}\textrm{ with } b = \alpha F^{T} R f + D^{T}_x \left( r d_x + \lambda_x \right) + D^{T}_y \left( r d_y + \lambda_y \right).
 \end{equation}
 Actually as $R$ is a row selector, $R^{T} R$ is a sparse diagonal matrix with non-zero entries on the selected Fourier coefficients and we cannot assure the invertibility of $H$. We find an approximate solution defining the positive definite matrix $ H_{\epsilon}~=~H~+~\epsilon I_n$ with small $\epsilon>0$ and solving the approximate system
\begin{equation}
H_{\epsilon}u = b + \epsilon \hat{u} \label{eq:linsystem},
\end{equation}
where we use the value of $u$ from the previous augmented Lagrangian iteration to estimate $\hat{u}=u^{l-1}$. In the resulting system, $H_{\epsilon}$ is block circulant and we can use the Fourier transform to decompose it as $H_{\epsilon}~=~\mathcal{F}^{T} C \mathcal{F}$, with $C~=~\alpha R^{T}R~+~r \mathcal{F}\left( D^{T}_x D_x~+~D^{T}_y D_y~\right) \mathcal{F}^{T}~+~\epsilon I_n$ a diagonal matrix. Consequently, the system \eqref{eq:linsystem} can easily be solved in the Fourier domain inverting the diagonal matrix $C$. In practice we use the FFT transform instead of doing the matrix multiplications with $\mathcal{F}$ and $\mathcal{F}^{T}$, which gives us a solution of complexity $\mathcal{O}(n\log n)$.

The minimization problem w.r.t. $\vect{d}$ corresponds to an $\ell_1$ - $\ell_2$ norm and can be solved by shrinkage.
We first note that \eqref{eq:min2} is equivalent to
\begin{align}
\min_{d_x, d_y} \sum_i \vert \vect{d}(i) \vert +  \frac{r}{2} \sum_i \vert \vect{d}(i) - \vect{z}(i)  \vert^2 ,\hspace{4pt}\textrm{with}\hspace{4pt} \vect{z} = \frac{1}{r}\vect{\lambda} - \nabla u  . \label{eq:en2}
\end{align}
The minimization of \eqref{eq:en2} can be done pixel-wise and the solution is given by the shrinkage operator $\mathcal{S}\left(\vect{z},1/r\right)$.
\begin{equation}
\vect{d}(i)=\max\Big\{\vert \vect{z}(i) \vert-\frac{1}{r},0\Big\} \frac{\vect{z}(i)}{\vert \vect{z}(i) \vert}  \hspace{8pt} i=1,\ldots, n
\end{equation}

\subsubsection{Regularization of normals} \label{sec:min_local2}
To regularize the normals at each iteration we have to solve
\begin{align}
\min_{\vert \vect{n}\vert \leq 1 }  \Vert \ \vert W \left( \nabla n_x, \nabla n_y \right) \vert \ \Vert_1 + \frac{\mu}{2} \Vert n_x - \hat{n}_x \Vert_2 ^2 + \frac{\mu}{2} \Vert n_y - \hat{n}_y \Vert_2 ^2 ,
\end{align}
where $W$ is a diagonal matrix with weights associated to weighted TV seminorm.
We use the same combination of splitting and augmented Lagrangian techniques than in Section \ref{sec:min_local1}. To avoid repetition, on the following we will simply write the form of the constraint minimization problem, the augmented Lagrangian and each of the subminimizations for a self-contained paper.

Equivalent constraint minimization is
\begin{align}
\min_{\substack{\vect{n}=\vect{m},  \vert \vect{m}\vert \leq 1 \\ \vect{d}=\nabla n_x, \vect{e} = \nabla n_y }}  \Vert \ \vert W \left(\vect{d}, \vect{e}\right) \vert \ \Vert_1  + \frac{\mu}{2} \Vert n_x - \hat{n}_x \Vert_2 ^2 + \frac{\mu}{2} \Vert n_y - \hat{n}_y \Vert_2 ^2 \label{eq:cmin2},
\end{align}
with associated augmented Lagrangian
\begin{align}
\mathcal{L}_2\left( \vect{n}, \vect{m}, \vect{d}, \vect{e} ,\vect{\lambda}, \vect{\nu} , \vect{\xi}\right) = \Vert \ \vert W \left(\vect{d}, \vect{e}\right) \vert \ \Vert_1 + \frac{\mu}{2} \Vert \vect{n} - \vect{\hat{n}} \Vert_2 ^2 + \nonumber \\ \lambda^{T}_{x} ( d_x - D_x n_x) + \lambda^{T}_{y} ( d_y - D_y n_x)   + \frac{r_d}{2} \Vert d_x - D_x n_x \Vert_2 ^2 +  \nonumber \\ \frac{r_d}{2} \Vert d_y - D_y n_y \Vert_2 ^2 + \nu^{T}_{x} ( e_x - D_x n_y) + \nu^{T}_{y} ( e_y - D_y n_y)   \nonumber \\ + \frac{r_e}{2} \Vert e_x - D_x n_y \Vert_2 ^2  + \frac{r_e}{2} \Vert e_y - D_y n_y \Vert_2 ^2 + \xi^{T}_x ( n_x - m_x )  \nonumber \\ + \frac{r_m}{2}\Vert n_x -m_x\Vert_2 ^2 + \xi^{T}_y ( n_y - m_y ) + \frac{r_m}{2}\Vert n_y - m_y\Vert_2 ^2
\label{eq:AL_local2}.
\end{align}

The resulting minimization method is presented in Algorithm~\ref{alg:AL_it2}.
\begin{algorithm}
\caption{Augmented Lagrangian method to solve \eqref{eq:cmin2}, regularizing normals}
\label{alg:AL_it2}
\begin{algorithmic}[1]
\State Initialize $ \vect{n}, \vect{m}, \vect{d}, \vect{e}, \vect{\lambda}, \vect{\nu}, \vect{\xi}$
\State For each iteration $l=1,2\ldots$, find an approximate minimizer of $\mathcal{L}_2$ with respect to variables $(n_x, n_y, \vect{m}, \vect{d}, \vect{e})$ with fixed Lagrange multipliers $\vect{\lambda}^l, \vect{\nu}^l, \vect{\xi}^l$:
\small{
\begin{align}
\vect{n}^{l} = \arg\min_{ \vect{n} } \mathcal{L}_2(\vect{n},  \vect{m}^{l-1}, \vect{d}^{l-1}, \vect{e}^{l-1}, \vect{\lambda}^{l-1}, \vect{\nu}^{l-1}, \vect{\xi}^{l-1} )\nonumber\\
\vect{m}^{l}= \arg\min_{ \vect{m} } \mathcal{L}(\vect{n}^{l},\vect{m},\vect{d}^{l-1}, \vect{e}^{l-1},\vect{\lambda}^{l-1},\vect{\nu}^{l-1}, \vect{\xi}^{l-1}  ) \nonumber\\
\vect{d}^{l}= \arg\min_{ \vect{d} } \mathcal{L}(\vect{n}^{l},\vect{m}^{l}, \vect{d}, \vect{e}^{l-1},\vect{\lambda}^{l-1},\vect{\nu}^{l-1}, \vect{\xi}^{l-1}  )\nonumber\\
\vect{e}^{l}= \arg\min_{ \vect{e} } \mathcal{L}(\vect{n}^{l},\vect{m}^{l} ,\vect{d}^{l}, \vect{e},\vect{\lambda}^{l-1},\vect{\nu}^{l-1}, \vect{\xi}^{l-1}  )\nonumber
\end{align}
}
\State Update Lagrange multipliers
\small{
\begin{align}
\lambda^{l}_x =& \lambda^{l-1}_x + r_d( d^{l}_x - D_x n^{l}_x ) \nonumber\\
\lambda^{l}_y =& \lambda^{l-1}_y + r_r( d^{l}_y  - D_y n^{l}_x )\nonumber\\
\nu^{l}_x =& \nu^{l-1}_x + r_e( e^{l}_x - D_x n^{l}_y ) \nonumber\\
\nu^{l}_y =& \nu^{l-1}_y + r_e( e^{l}_y  - D_y n^{l}_y )\nonumber\\
\xi^{l} =& \xi^{l-1} + r_m( \vect{n} - \vect{m} ) \nonumber
\end{align}
}
\State Stop the iterative process when $\frac{ \Vert \vect{n}^l - \vect{n}^{l-1} \Vert}{\Vert \vect{n}^l \Vert} <\epsilon$.
\end{algorithmic}
\end{algorithm}

The sub-minimization problem with respect to $n_x$ can be written as follows:
\begin{align}
\min_{n_x} \ \frac{\mu}{2} \Vert n_x - \hat{n}_x \Vert_2 ^2 +  \xi^{T}_x ( n_x - m_x ) + \frac{r_m}{2}\Vert n_x - m_x\Vert_2 ^2 \nonumber \\
+  \frac{r}{2} \Vert d_x + \frac{1}{r_d}\lambda_x - D_x n_x \Vert_2 ^2 + \frac{r_d}{2} \Vert d_y + \frac{1}{r_d}\lambda_y - D_y n_x  - \Vert_2 ^2
\end{align}
We see that it reduces to a quadratic minimization, with positive definite Hessian $ H=(\mu + r_m) I_n+r_d D^{T}_x D_x+r_d D^{T}_y D_y$.
The optimality conditions read 
\begin{equation}
 H n_x = \mu \hat{n}_x + r_m m_x + D^{T}_x \left( r_d d_x + \lambda_x \right) + D^{T}_y \left( r_d d_y + \lambda_y \right) - \xi_x.
 \end{equation}
As before, the resulting $H$ is block circulant and we can use the Fourier transform to decompose it as $H~=~\mathcal{F}^{T} C \mathcal{F}$, with $C~=~(\mu~+~r_m) I_n~+~r_d \mathcal{F}\left( D^{T}_x D_x~+~D^{T}_y D_y \right) \mathcal{F}^{T} $ a diagonal matrix. We solve the linear system in the Fourier domain efficiently with the FFT transform. Observe that the minimization problem with respect to $n_y$ has the same form and can be solved with the same technique.

The minimization problem w.r.t. $\vect{d}$ corresponds to the $\ell_1$ - $\ell_2$ problem
\begin{align}
\min_{d_x, d_y} \sum_i \vert w(i)\vect{d}(i) \vert +  \frac{r_d}{2} \sum_i \vert \vect{d}(i) - \vect{z}(i)  \vert^2 ,\hspace{4pt}\textrm{with}\hspace{4pt} \vect{z} = \frac{1}{r_d}\vect{\lambda} - \nabla n_x  ,
\end{align}
which is equivalent to the following minimization (note $w(i)>0$)
\begin{align}
\min_{d_x, d_y} \sum_i \vert \vect{d}(i) \vert +  \frac{r_d}{2w(i)} \sum_i \vert \vect{d}(i) - \vect{z}(i)  \vert^2 ,\hspace{4pt}\textrm{with}\hspace{4pt} \vect{z} = \frac{1}{r_d}\vect{\lambda} - \nabla n_x .
\end{align}
A similar problem has already been solved in Section \ref{sec:min_local1} with the shrinkage operator, which is now adapted to include the weights $w$. The solution is then
\begin{equation}
\vect{d}(i)=\max\Big\{\vert \vect{z}(i) \vert-\frac{w(i)}{r},0\Big\} \frac{\vect{z}(i)}{\vert \vect{z}(i) \vert}  \hspace{8pt} i=1,\ldots, n
\end{equation}
\\
Due to the symmetry of the problems, the same minimization technique is used for $\vect{e}$.

Finally, the minimization problem w.r.t. $\vect{m}$ reads
\begin{align}
\min_{ \vert \vect{m} (i) \vert = 1 } \frac{r_m}{2} \sum_i \vert \vect{m}(i) - \vect{z}(i)  \vert^2 ,\hspace{4pt}\textrm{with}\hspace{4pt} \vect{z} = \vect{n} + \frac{1}{r_m}\vect{\xi} 
\end{align}
and can be solved pixel-wise. For each pixel we have the following 2-D problem: given a point in space with coordinates $\vect{z}(i)\in\mathbb{R}^2$ we want to find the point in the unit ball minimizing its distance to $\vect{z}(i)$. It is clear that the solution corresponds to the projection of the unconstrained minimizer $\vect{z}(i)$ into the unit ball, i.e 
\begin{align}
\vect{m}(i) = \begin{cases} \vect{z}(i) & \vert \vect{z}(i)\vert \leq 1 \\ \frac{\vect{z}(i)}{\vert \vect{z}(i) \vert} & \textrm{otherwise}\end{cases} .
\end{align}

\subsection{Minimizations of non local normal-guided CS}\label{sec:min_NL}

In the discrete setting, the NL gradient is a linear operator.  Arranging the image as a column vectors, it can be computed as a sparse matrix multiplication $\nabla_{G} u = D u$. The matrix $D\in \mathbb{R}^{N\times n}$ ($N = \vert E \vert$ indicates the number of nodes in the graph) is derived from the weights associated to the edges and is usually sparse. Consequently $d = D u \in \mathbb{R}^{N}$ is also a vector column, with as many components associated to node $i$ as neighbours this node has. With the vector notation, the inner product of two vectors fields $\vect{d},\vect{e}$ defined in G is then computed as $\ipG{ \vect{d}}{\vect{e}} = \vect{d}^{T}\vect{e}$. 
As in the continuous setting, the NL divergence $\Div_{G}$ is derived from its adjoint relation with the NL gradient $\nabla_{G}^{*} = -\Div_{G}$ and, consequently, in matrix notation it corresponds to $\Div_{G} \vect{d} = -D^{T} \vect{d}$.

Since the minimization associated to \eqref{eq:ROF} has already been explained for the vectorial case, in the next paragraphs we focus on the minimizations associated to non local operators \eqref{eq:NLCSnormal} and \eqref{eq:NLROF}.

\subsubsection{Minimization associated to CS reconstruction matching non local normals} \label{sec:min_NL2}

With the previous notation, the minimization problem \eqref{eq:NLCSnormal} reads
\begin{align}
u = \arg \min_{u} \ \Vert \ \vert Du \vert_G \ \Vert_1  +\gamma v^{T} u  +  \frac{\alpha}{2} \Vert Au - f \Vert_2 ^2 \hspace{4pt}
\label{eq:minNL2}
\end{align}
with $v = \Div_G \vect{n}_{G}$. This minimization is also reformulated as a constraint minimization and solved efficiently with augmented Lagrangians. Compared to the local minimizations, in the splitting step we require an additional variable $s$ to have efficient and analytic solutions for the posterior subminimization problems. The resulting constraint minimization formulation of \eqref{eq:minNL2} is 
\begin{equation}
\min_{u,s,\vect{d}} \ = \ \Vert \ \vert \vect{d} \vert_G \ \Vert_1 +  v^{T}u +  \frac{\alpha}{2} \Vert Au - f \Vert_2 ^2 \ \textrm{ s.t. }\ \left\{
  \begin{array}{ l}
\vect{d}= D u\\
s =  u 
\end{array} \right.
\label{eq:NLcmin2}
\end{equation}
\\
The Lagrangian in that case reads
\begin{align}
\mathcal{L}_3\left( u, s, \vect{d}, \vect{\lambda_d}, \lambda_u \right) = \ \Vert \ \vert \vect{d} \vert_G \ \Vert_1 +  v^{T}u +  \frac{\alpha}{2} \Vert Au - f \Vert ^2  \nonumber \\+ \vect{\lambda_d}^{T} \left( \vect{d} - D u\right) + \frac{r_d}{2} \Vert \vect{d} - D u \Vert^2 + \lambda_u^{T}\left( u - s\right) + \frac{r_u}{2} \Vert u - s \Vert^2 .
\label{eq:NLAL2}
\end{align}

The resulting minimization method is presented in Algorithm~\ref{alg:AL_it3}, where we have also hinted the solution to each of the subminimization problems.
\begin{algorithm}
\caption{Augmented Lagrangian method for CS reconstruction matching normals by \eqref{eq:NLcmin2}}
\label{alg:AL_it3}
\begin{algorithmic}[1]
\State Initialize $ u, s, \vect{d}, \vect{\lambda_d}, \lambda_u$
\State For each iteration $l=1,2\ldots$, find an approximate minimizer of $\mathcal{L}_3$ with respect to variables $(u, s, \vect{d})$ with fixed Lagrange multipliers $\vect{\lambda_d}^{l}, \lambda^{l}_u$:
\small{
\begin{align}
u = \arg\min_{ u } \mathcal{L}_3(u, s^{l-1},  \vect{d}^{l-1}, \vect{\lambda_d}^{l-1}, \lambda^{l-1}_u ) & \textrm{ solved with conjugate gradients}\nonumber\\
s= \arg\min_{ s } \mathcal{L}_3(u^{l}, s,  \vect{d}^{l-1}, \vect{\lambda_d}^{l-1}, \lambda^{l-1}_u ) & \textrm{ solved in Fourier domain}\nonumber\\
\vect{d}= \arg\min_{ \vect{d} } \mathcal{L}_3(u^{l}, s^{l},  \vect{d}, \vect{\lambda_d}^{l-1}, \lambda^{l-1}_u )& \textrm{ solved by non local shrinkage} \nonumber
\end{align}
}
\State Update Lagrange multipliers
\small{
\begin{align}
\vect{\lambda_d}^{l} =& \vect{\lambda_d}^{l-1} + r_d( \vect{d}^{l} - D u^{l} ) \nonumber\\
\lambda^{l}_u =& \lambda^{l-1}_u + r_u( u^{l}  - s^{l} )\nonumber
\end{align}
}
\State Stop the iterative process when $\frac{ \Vert u^l - u^{l-1} \Vert}{\Vert u^l \Vert} <\epsilon$.
\end{algorithmic}
\end{algorithm}

The minimization w.r.t $u$ corresponds to the following quadratic positive definite problem
\begin{equation}
\min_{u} \ v^{T}u + \vect{\lambda_d}^{T} \left( \vect{d} - D u\right) + \frac{r_d}{2} \Vert \vect{d} - D u \Vert^2 + \lambda_u^{T}\left( u - s\right) \nonumber\\+ \frac{r_u}{2} \Vert u - s \Vert^2 .
\label{eq:NLminu}
\end{equation}
We find its minimizer solving its optimality conditions, which provide the following system of linear equations 
\begin{equation}
\left( r_u I + r_d D^{T}D\right) u = -\gamma v - \lambda_u + r_u s + D^{T}\left( \vect{\lambda_d} + r_d \vect{d}\right).
\label{eq:NLminu_opt}
\end{equation}
Matrix $K = r_u + r_d D^{T}D$ is sparse, symmetric and positive definite and we have efficient algorithms to invert it. We choose an iterative method to invert the matrix, initializing it from the previous solution to the minimization problem $u^{l-1}$. In particular we use the conjugate gradient method to exploit the symmetry and positive definition of $K$, with preconditioning matrix given by its incomplete Cholesky factorization. The resulting method is very fast, converging to enough precision with $2-3$ iterations of the conjugate gradient method.

The minimization w.r.t $s$ is also a quadratic problem which can be efficiently solved, in that case in the Fourier domain.
The problem reads
\begin{equation}
\min_{s} \  \frac{\alpha}{2} \Vert Au - f \Vert ^2 + \lambda_u^{T}\left( u - s\right) + \frac{r_u}{2} \Vert u - s \Vert^2 .
\label{eq:NLmins}
\end{equation}
The optimality conditions in that case are 
\begin{align}
\left( \alpha A^{T} A + r_u I_n\right) s = \alpha A^{T}f + \lambda_u + r_u u.
\label{eq:NLmins_opt}
\end{align}
As before, the matrix $A^{T} A + r_u I_n = F^{T}CF$ is block-circulant and the resulting system is diagonal in the Fourier domain with $C=R^{T}R +  r_u I_n$. Therefore \eqref{eq:NLmins_opt} can be efficiently solved with the FFT.

The minimization with respect to $d$ is equivalent to
\begin{equation}
\min_{\vect{d}} \ = \ \Vert \ \vert \vect{d} \vert_G \ \Vert_1 +  \frac{r_d}{2} \Vert \vect{d} -  \vect{z} \Vert^2  \ \textrm{ with  }\ \vect{z} = D u - \frac{\vect{\lambda_d}}{r_d} \label{eq:NLmind}
\end{equation}
As in the local case, this minimization is decoupled for each pixel $i$ as follows
\begin{equation}
\min_{d(i,j)\ j \sim i } \ = \ \sqrt{ \sum_{j \sim i} d^2(i,j) } +  \sum_{j \sim i}  \frac{r_d}{2} \left( d(i,j)-  z(i,j) \right)^2  
\end{equation}
and can be solved by a straight-forward extension of the shrinkage operator to the graph. That is, for each node neighbour to $i$ the solution is given by
\begin{equation}
\vect{d}^\star(i,j)=\max\Big\{ \vert z \vert_G(i) -\frac{1}{r_d},0\Big\} \frac{z(i,j)}{ \vert z \vert_G (i)} . 
\end{equation}

\subsubsection{Minimization associated to the regularization of  normals} \label{sec:min_NL1}

With the previous notation, the minimization problem \eqref{eq:NLROF} reads
\begin{align}
v =  \arg \min_{v}  \ \Vert \ \vert Dv \vert_G \ \Vert_1 + \frac{\mu}{2}\Vert v - \hat{v} \Vert_2 ^2 \label{eq:minNL1}
\end{align}
As in the local case, we decouple the $\ell_1$ and $\ell_2$ problems defining an additional variable $\vect{d} = Du$ and rewrite \eqref{eq:minNL1} as the following constraint minimization problem
\begin{align}
\min_{v,\vect{d}} \ = \ \Vert \ \vert \vect{d} \vert_G \ \Vert_1 +  \frac{\mu}{2} \Vert v - \hat{v} \Vert_2 ^2 \hspace{4pt} \textrm{ s.t. } \hspace{4pt} \vect{d}= D u
\label{eq:NLcmin1}
\end{align}
with associated augmented Lagrangian 
\begin{align}
\mathcal{L}_4\left( u, \vect{d}, \vect{\lambda_d} \right) = \ \Vert \ \vert \vect{d} \vert_G \ \Vert_1 +  \frac{\mu}{2} \Vert v - \hat{v} \Vert_2 ^2  \nonumber \\+ \vect{\lambda_d}^{T} \left( \vect{d} - D v\right) + \frac{r_d}{2} \Vert \vect{d} - D v \Vert_2 ^2 .
\label{eq:NLAL1}
\end{align}

To minimize the Lagrangian $\mathcal{L}_3$ with respect to $u,\vect{d}$, we alternate the direction of minimization with respect to each variable and proceed as indicated by Algorithm~\ref{alg:AL_it3}, where we have also hinted the solution to each of the subminimization problems.
\begin{algorithm}
\caption{Augmented Lagrangian method to regularize non local divergence of normals from \eqref{eq:NLcmin1}}
\label{alg:AL_it4}
\begin{algorithmic}[1]
\State Initialize $ u, \vect{d}, \vect{\lambda_d}$
\State For each iteration $l=1,2\ldots$, find an approximate minimizer of $\mathcal{L}_4$ with respect to variables $(u, \vect{d})$ with fixed Lagrange multipliers $\vect{\lambda_d}^{l}$:
\small{
\begin{align}
v = \arg\min_{ v } \mathcal{L}_4(v, \vect{d}^{l-1}, \vect{\lambda_d}^{l-1} ) & \textrm{ solved with conjugate gradient}\nonumber\\
\vect{d}= \arg\min_{ \vect{d} } \mathcal{L}_4(v^{l},\vect{d}, \vect{\lambda_d}^{l-1} )& \textrm{ solved by non local shrinkage} \nonumber
\end{align}
}
\State Update Lagrange multipliers
\small{
\begin{align}
\vect{\lambda_d}^{l} =& \vect{\lambda_d}^{l-1} + r_d( \vect{d}^{l} - D u^{l} ) \nonumber
\end{align}
}
\State Stop the iterative process when $\frac{ \Vert v^l - v^{l-1} \Vert}{\Vert v^l \Vert} <\epsilon$.
\end{algorithmic}
\end{algorithm}

The minimization w.r.t $v$ corresponds to the following quadratic positive definite problem
\begin{equation}
\min_{v} \  \frac{\mu}{2} \Vert v - \hat{v} \Vert_2 ^2 + \vect{\lambda_d}^{T} \left( \vect{d} - D v\right) + \frac{r_d}{2} \Vert \vect{d} - D v \Vert_2 ^2 .
\label{eq:NLminv}
\end{equation}
We find its minimizer by solving its optimality conditions, which provide the following system of linear equations 
\begin{equation}
\left( \mu I+ r_d D^{T}D\right) v = \mu \hat{v} + D^{T}\left( \vect{\lambda_d} + r_d \vect{d}\right).
\label{eq:NLminv_opt}
\end{equation}
We find the same form of matrix $K = \mu I + r_d D^{T}D$ than in \eqref{eq:NLminu_opt} and, therefore, we solve with linear system \eqref{eq:NLminu_opt} with the same conjugate gradient method.

The minimization with respect to $d$ is equivalent to \eqref{eq:NLmind} changing $u$ for $v$, in particular we have 
\begin{equation}
\min_{\vect{d}} \ = \ \Vert \ \vert \vect{d} \vert_G \ \Vert_1 +  \frac{r_d}{2} \Vert \vect{d} -  \vect{z} \Vert^2  \ \textrm{ with  }\ \vect{z} = D u - \frac{\vect{\lambda_d}}{r_d}
\end{equation}
and is solved with the same adaptation of the shrinkage operator to the graph. For each node neighbour to $i$, the solution is given by
\begin{equation}
d^\star(i,j)=\max\Big\{ \vert z \vert_G(i) -\frac{1}{r_d},0\Big\} \frac{z(i,j)}{ \vert z \vert_G (i)} . 
\end{equation}

\section{Numerical results} \label{sec:experiments}
In this section we present some of the numerical results obtained with our method and compare it to other techniques. We compare the local version of our method to standard CS recovery algorithm \eqref{eq:CSmin} and to the edge-guided CS proposed in \cite{Guo2010}. The non local version of our model is compared to the non local CS recovery \eqref{eq:NLCSmin}, which does not take into account the geometric information of the non local gradients into the recovery process. For the non local case, in our model we have regularized the divergence of the normal with the standard ROF model.

We use partial Fourier measurements for our reconstruction and perform radial sampling on $R$ with different number of measurements in relation to the size of the signal (we specify it with the ratio $\frac{m}{n}$). For a fair comparison, we have used the same robust edge detector \eqref{eq:edge_detector} for the edge-guided CS and our method and we have implemented the minimizations with the same splitting and augmented Lagrangian techniques for all the methods. The parameter $\alpha$, which is related to the noise present in the CS measurements, has been manually tuned to obtain best reconstruction with the standard CS recovery models \eqref{eq:CSmin} and \eqref{eq:NLCSmin} and used with the other methods. The other parameters of our model $\gamma,\mu$ have also been chosen manually to obtain good CS recovery in terms of SNR. We have observed that $\gamma$ (which controls the weight given to the alignment of the normals) takes similar values for the same kind of images (textured or brain IRM images) and remains stable for different sparsity and noise levels. On the other hand, the parameter $\mu$ controlling the smoothness of the estimated normals decreases when the number of measurements decreases or the noise level increases because the partial reconstructions and the estimated normals are noisier and require more regularization.

\begin{figure*}[h]
\centering
\subfigure[Original image]{\includegraphics[width=8cm]{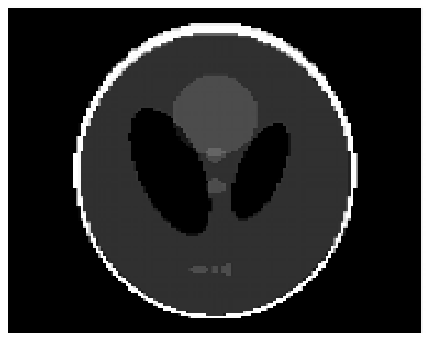}}
\subfigure[Back-projection, $2.77$ dB]{\includegraphics[width=8cm]{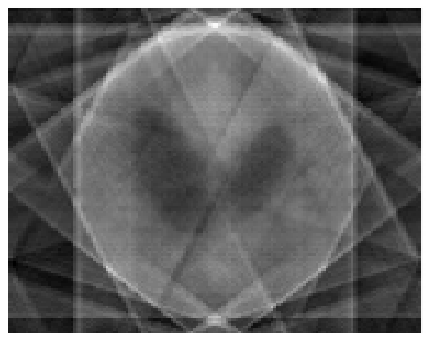}}
\\
\centering
\subfigure[TV reconstruction, $7.33$ dB]{\includegraphics[width=8cm]{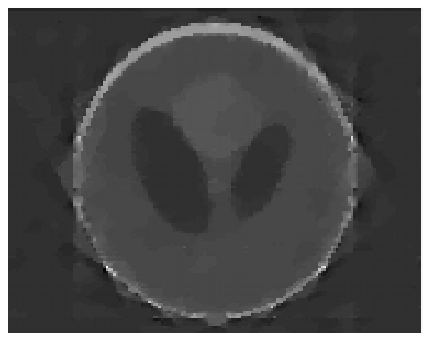}}
\subfigure[non local TV reconstruction, $28.28$ dB]{\includegraphics[width=8cm]{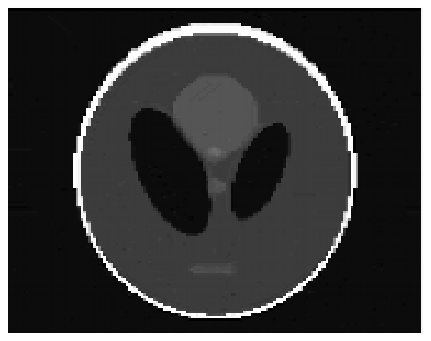}}
\\
\centering
\subfigure[proposed local method, $12.78$ dB]{\includegraphics[width=8cm]{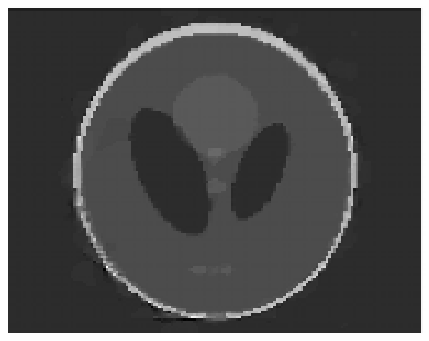}}
\subfigure[proposed non local method, $31.26$ dB]{\includegraphics[width=8cm]{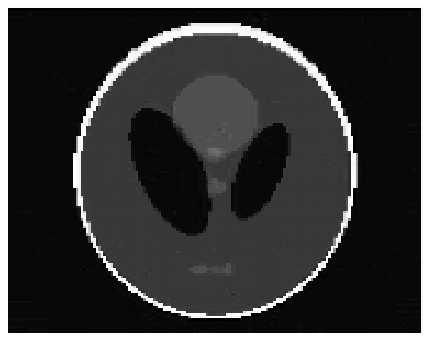}}
\caption{Reconstruction of Shepp-Logan phantom from 8\% of measurements in Fourier domain.
}
\label{fig:phantom8}
\end{figure*}

\begin{figure*}[h]
\centering
\subfigure[Original image]{\includegraphics[width=8cm]{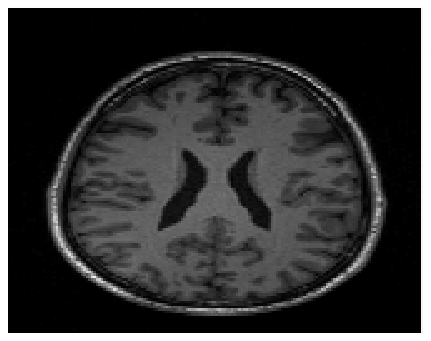}}
\subfigure[Back-projection, $9.25$ dB]{\includegraphics[width=8cm]{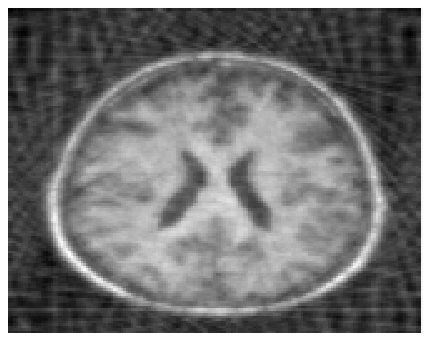}}
\\
\centering
\subfigure[TV reconstruction, $17.14$ dB]{\includegraphics[width=8cm]{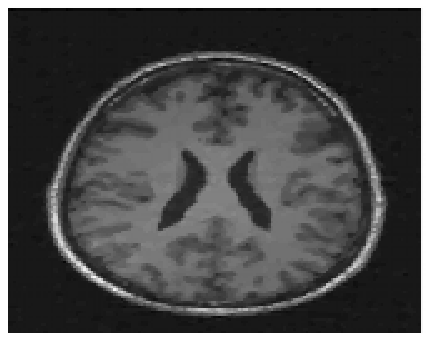}}
\subfigure[non local TV reconstruction, $18.96$ dB]{\includegraphics[width=8cm]{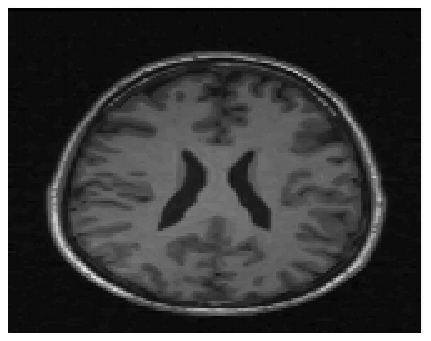}}
\\
\centering
\subfigure[proposed local method, $18.56$ dB]{\includegraphics[width=8cm]{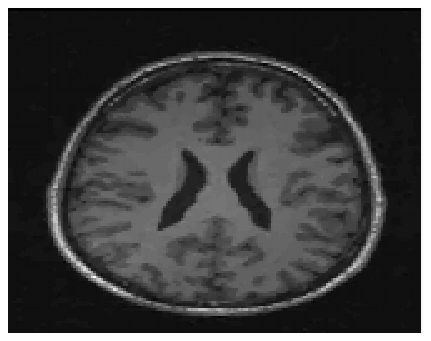}}
\subfigure[proposed non local method, $20.39$ dB]{\includegraphics[width=8cm]{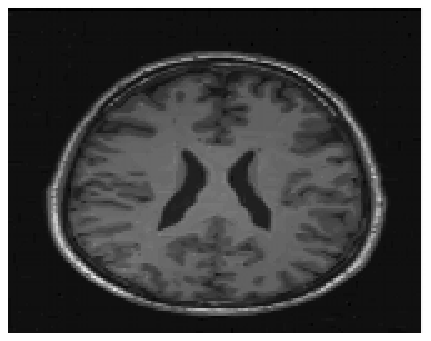}}
\caption{Reconstruction of MRI brain image from 12\% of measurements in Fourier domain. 
 }
\label{fig:slice12}
\end{figure*}

\begin{figure*}[h]
\centering
\subfigure[Original image]{\includegraphics[width=8cm]{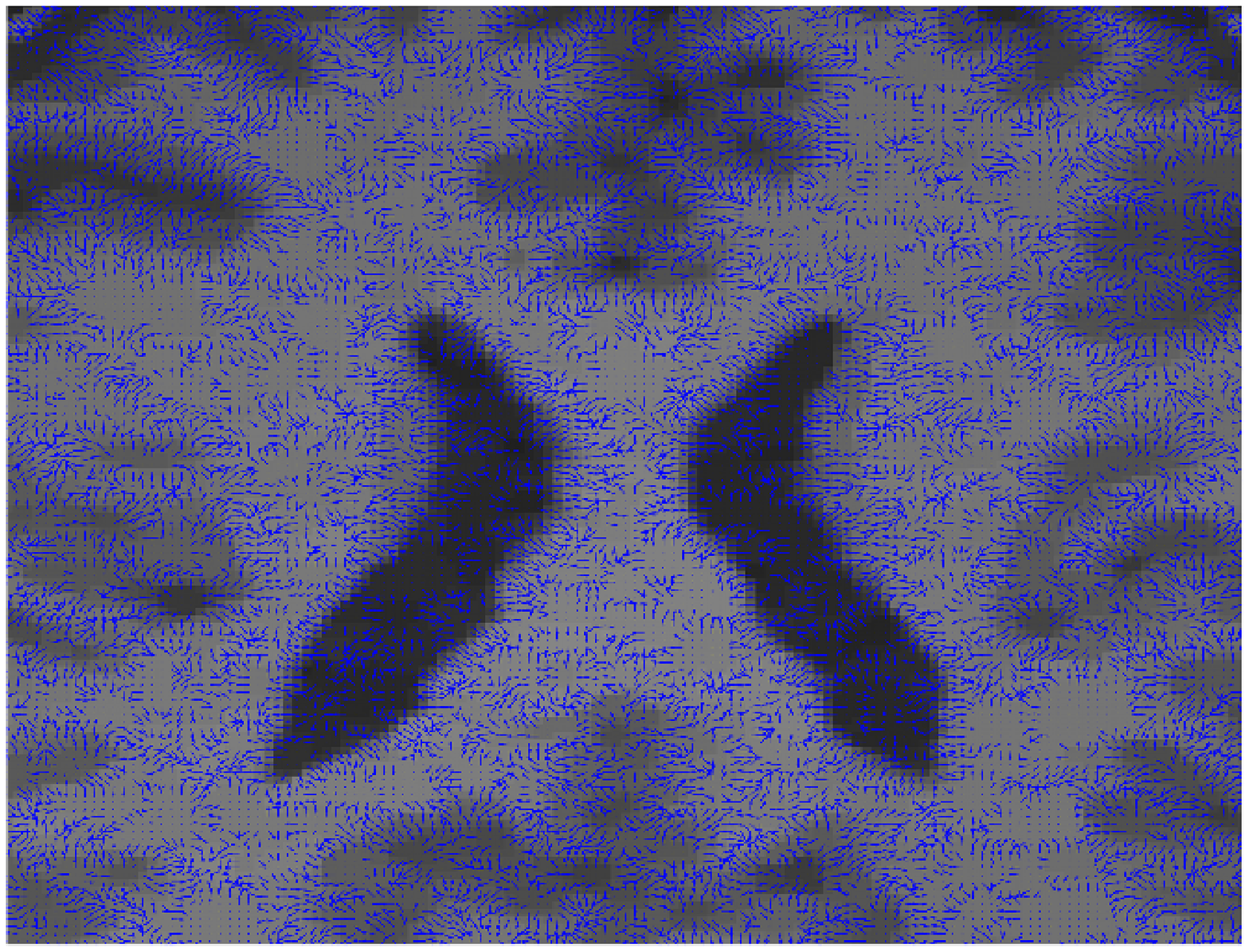}}
\subfigure[Proposed local technique]{\includegraphics[width=8cm]{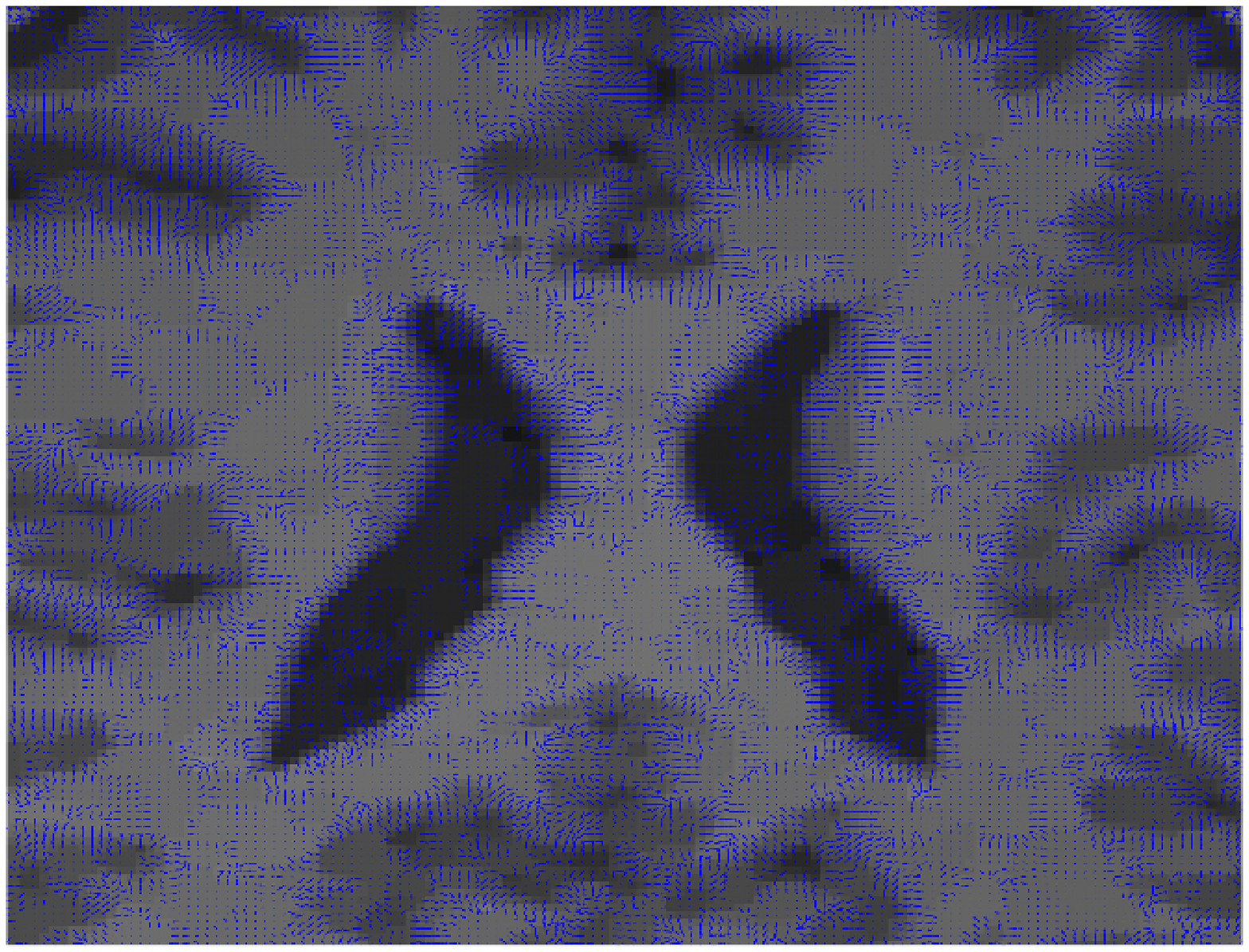}}
\caption{Zoom on reconstructed brain MRI image from 12\% of measurements in Fourier domain. We superpose the reconstructed signals with the normals associated to their level sets for the standard TV solution (left) and for the local version of our method (right). Our method is able to better reconstruct the normals and shapes of the image. }
\label{fig:zoom_normal_MRI}
\end{figure*}

In a first set of experiments we test our method with MRI images, first with the Shepp-Logan phantom from Figure \ref{fig:phantom8} and then with a real MRI brain image from Figure \ref{fig:slice12}; where we include also the results with back-projection, i.e. the solution to $f=Au$ with smallest $l_2$ norm. 
\\
Table \ref{tab:MRI_clean} show the quantitative results of the  different CS reconstruction methods for MRI images. Our method always outperforms the standard TV reconstruction and the edge-guided CS technique. In the experiments, both the edge-guided CS and our proposed method are initialized with the TV solution and, therefore, always improve its reconstruction. Comparing the gains of these two methods with respect to the TV reconstruction, we observe that our method more than doubles the gain of edge-guided CS. Figures\ref{fig:phantom8}-\ref{fig:zoom_normal_MRI} show qualitatively the improvement of our method over TV reconstruction. In the case of the phantom we are able to better reconstruct the phantom with fewer measurements both in the local and non local case, while with a real MRI image our reconstruction is able to capture better non-dominant edges of the white-grey matter interface. In Figure \ref{fig:zoom_normal_MRI} we explicitly compare the normals associated to the TV solution and the regularized normals of our local reconstruction for the real brain MRI image. We observe that our method is able to better reconstruct the normals, and therefore the shapes, of the image (to clearly see the differences please check the PDF, not the print out). 
\\
Performance improves with non local regularization, with our method outperforming the non local CS reconstruction for all the experiments. As expected, the gain of our method compared to TV is lower than in the local approach because we loose part of the directional information of the normals by denoising their divergence instead of the vector fields. 
\begin{table}[h]
\centering
\tabcolsep 3pt
 \begin{tabular}{|l | c |c c c| c c|}
 \hline
Image & $\frac{m}{n}$ & \multicolumn{3}{c|}{local CS} & \multicolumn{2}{c|}{ non-local CS }\\
& & TV & edge CS  & normal CS & TV & normal CS \\
\hline   
Phantom & 8\% &  7.33 dB & 7.37 dB & \textbf{12.78 dB} & 28.28 dB & \textbf{33.13 dB} \\
Phantom & 12\% & 38.60 dB & 45.33 dB & \textbf{56.14 dB} & 61.84 dB & \textbf{74.57 dB} \\
Brain & 12\% & 17.14 dB & 17.38 dB & \textbf{17.71 dB} & 18.96 dB & \textbf{20.39 dB} \\ 
Brain & 20\% & 22.16 dB & 22.35 dB & \textbf{23.82 dB} &  23.13 dB & \textbf{24.12 dB} \\ 
\hline
\end{tabular} 
\caption{Comparison of CS reconstruction for MRI images. The first three columns show the results with the standard TV term in the regularization: TV stands for the model \eqref{eq:CSmin}, edge CS for \eqref{eq:edgeCS} and normal CS for our method \eqref{eq:CSnormal}. The last two columns correspond to the definition of NL TV: NL-TV corresponds to the standard non local CS recovery \eqref{eq:NLCSmin} and NL normal CS for the proposed non local method \eqref{eq:NLCSnormal}. }
\label{tab:MRI_clean}
\end{table}
For each image we also added different levels of Gaussian noise ($\sigma_n$) to the signal to investigate the robustness of our method to noise. Results are shown in table \ref{tab:MRI_noisy}. We observe that we are more robust to noise than edge-guided CS (which in fact does not improve the TV reconstruction for noise levels $\sigma_n = 15\%$, $\sigma_n = 10\%$) thanks to regularization step on the estimation of the normals. As before, non local regularization improves CS reconstruction, we observe that our non local method outperforms again the non local TV and is also robust to noise.  
\begin{table}[h]
\centering
\tabcolsep 3pt
 \begin{tabular}{|l | c |c c c| c c|}
 \hline
Image & noise & \multicolumn{3}{c|}{local CS} & \multicolumn{2}{c|}{ non-local CS }\\
$\frac{m}{n}=12\%$ &$\sigma_n$ & TV & edge CS  & normal CS & TV & normal CS \\
\hline   
 & 5\% &  11.90 dB & 11.91 dB & \textbf{12.90 dB} & 17.92 dB & \textbf{18.36 dB} \\
Phantom & 10\% & 8.37 dB & 8.38 dB & \textbf{9.44 dB} & 12.15 dB & \textbf{13.03 dB} \\
 & 15\% &  6.59 dB & 6.59 dB & \textbf{7.28 dB} & 10.09 dB & \textbf{10.27 dB} \\
 \hline
 & 5\% & 13.37 dB & 13.36 dB & \textbf{13.78 dB} & 14.86 dB & \textbf{15.00 dB} \\ 
Brain & 10\% & 10.88 dB & 10.88 dB & \textbf{11.57 dB} & 12.31 dB & \textbf{12.50 dB} \\ 
 & 15\% &  9.89 dB & 9.89 dB & \textbf{10.48 dB} & 10.94 dB & \textbf{11.19 dB} \\
\hline
\end{tabular} 
\caption{Comparison of CS reconstruction for noisy MRI images with $12\%$ of samples and different levels $\sigma_n$ of gaussian noise. The first three columns show the results with the standard TV term in the regularization: TV stands for the model \eqref{eq:CSmin}, edge CS for \eqref{eq:edgeCS} and normal CS for our method \eqref{eq:CSnormal}. The last two columns correspond to the definition of NL TV: NL-TV corresponds to the standard non local CS recovery \eqref{eq:NLCSmin} and NL normal CS for the proposed non local method \eqref{eq:NLCSnormal}.  }
\label{tab:MRI_noisy}
\end{table}

\begin{figure*}[h]
\centering
\subfigure[Original image]{\includegraphics[width=8cm]{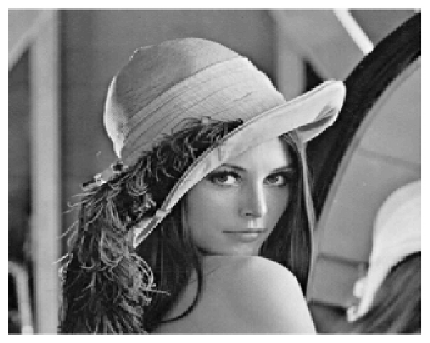}}
\subfigure[Back-projection, $8.12$ dB]{\includegraphics[width=8cm]{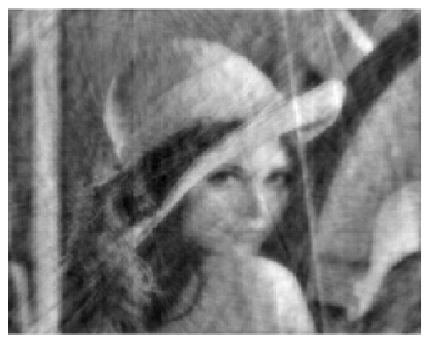}}
\\
\centering
\subfigure[TV reconstruction, $14.53$ dB]{\includegraphics[width=8cm]{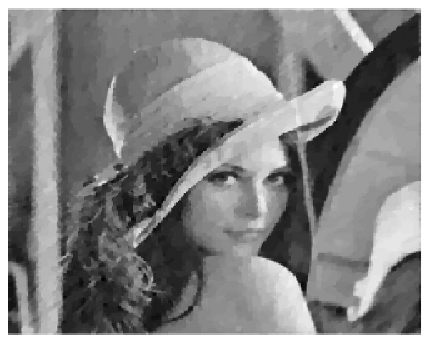}}
\subfigure[non local TV reconstruction, $15.82$ dB]{\includegraphics[width=8cm]{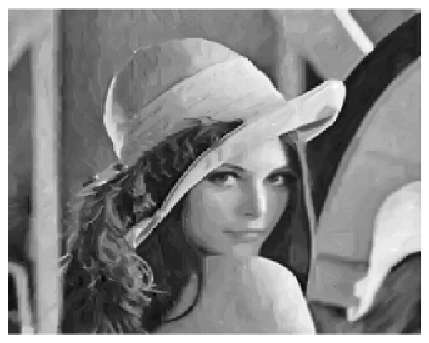}}
\\
\centering
\subfigure[proposed local method, $14.86$ dB]{\includegraphics[width=8cm]{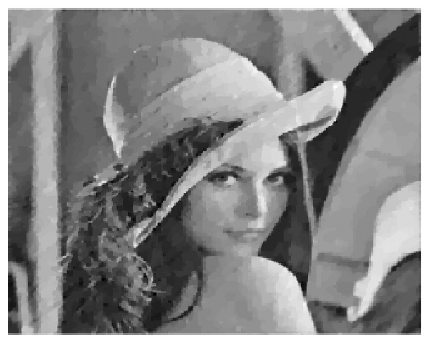}}
\subfigure[proposed non local method, $16.79$ dB]{\includegraphics[width=8cm]{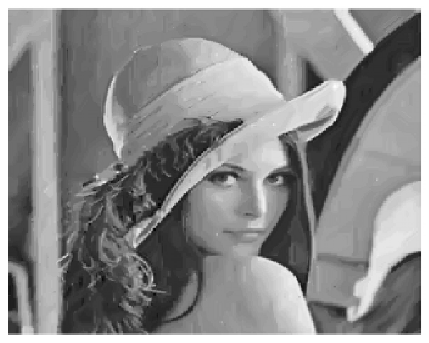}}
\caption{Reconstruction of Lena from 12\% of measurements in Fourier domain. }
\label{fig:lena12}
\end{figure*}

\begin{figure*}[h]
\centering
\subfigure[Original image]{\includegraphics[width=8cm]{lena.eps}}
\subfigure[Back-projection, $9.71$ dB]{\includegraphics[width=8cm]{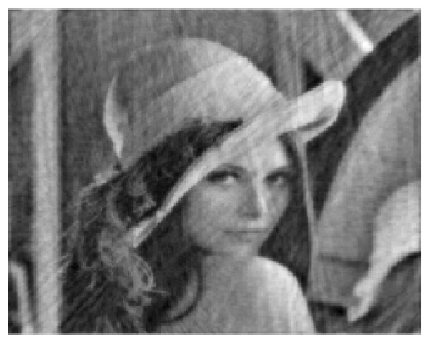}}
\\
\centering
\subfigure[TV reconstruction, $18.44$ dB]{\includegraphics[width=8cm]{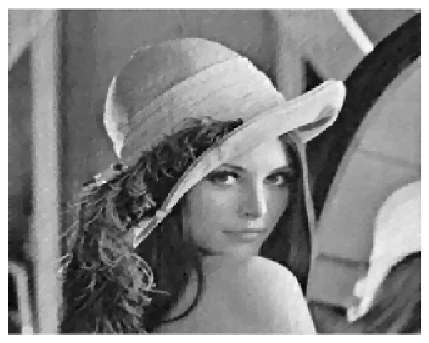}}
\subfigure[non local TV reconstruction, $19.45$ dB]{\includegraphics[width=8cm]{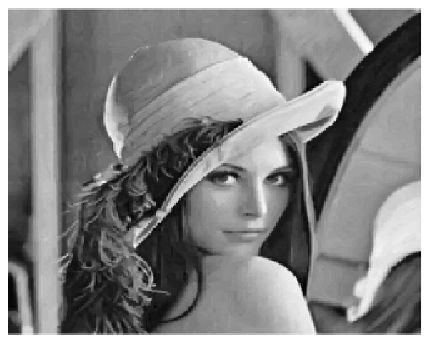}}
\\
\centering
\subfigure[proposed local method, $19.27$ dB]{\includegraphics[width=8cm]{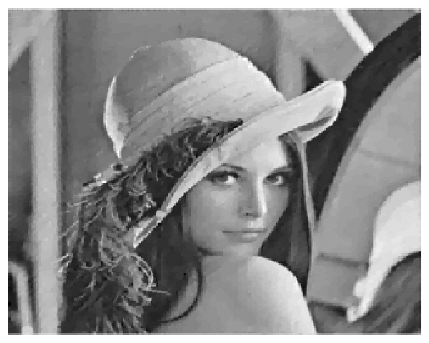}}
\subfigure[proposed non local method, $21.09$ dB]{\includegraphics[width=8cm]{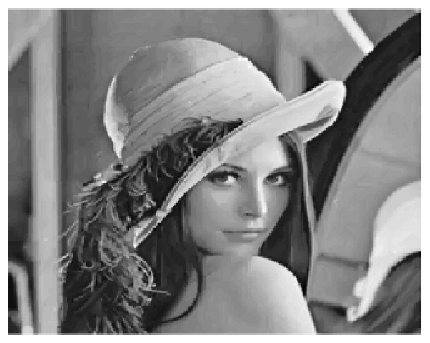}}
\caption{Reconstruction of Lena from 20\% of measurements in Fourier domain.}
\label{fig:lena20}
\end{figure*}

\begin{figure*}[h]
\centering
\subfigure[Original image]{\includegraphics[width=8cm]{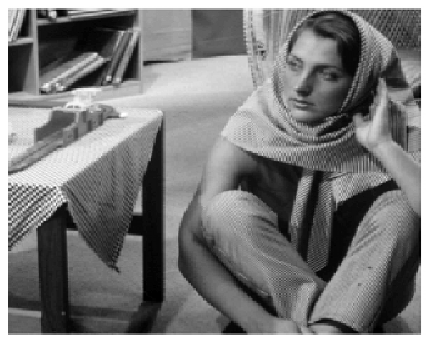}}
\subfigure[Back-projection, $8.30$ dB]{\includegraphics[width=8cm]{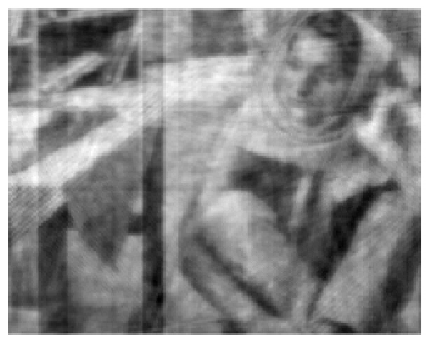}}
\\
\centering
\subfigure[proposed local method, $13.59$ dB]{\includegraphics[width=8cm]{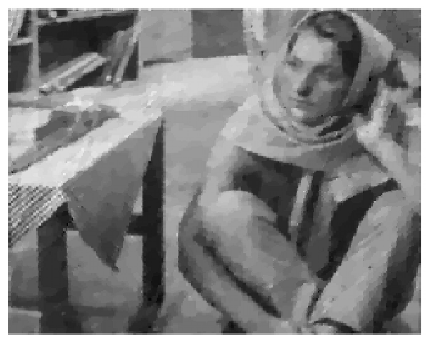}}
\subfigure[proposed non local method, $15.52$ dB]{\includegraphics[width=8cm]{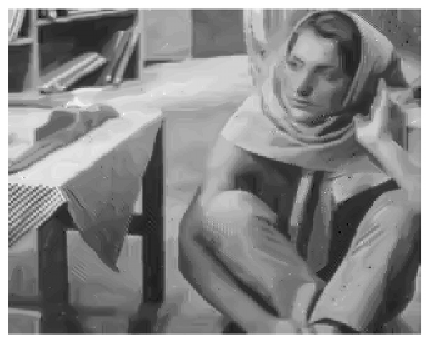}}
\caption{Reconstruction of Barbara from 12\% of measurements in Fourier domain. }
\label{fig:barbara12}
\end{figure*}

\begin{figure*}[h]
\centering
\subfigure[Original image]{\includegraphics[width=8cm]{barbara.eps}}
\subfigure[Back-projection, $10.07$ dB]{\includegraphics[width=8cm]{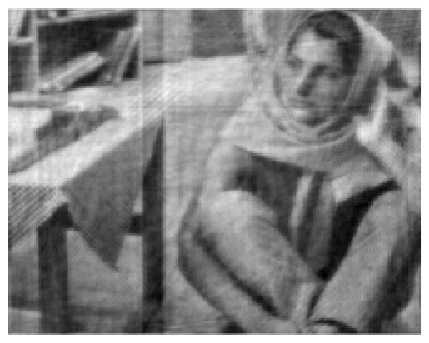}}
\\
\centering
\subfigure[proposed local method, $17.13$ dB]{\includegraphics[width=8cm]{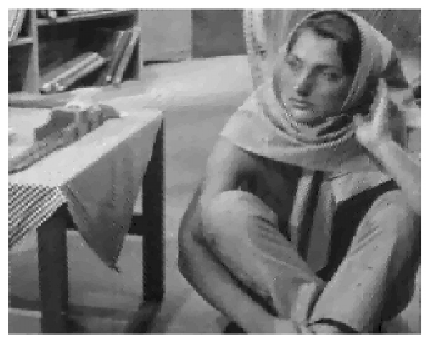}}
\subfigure[proposed non local method, $18.92$ dB]{\includegraphics[width=8cm]{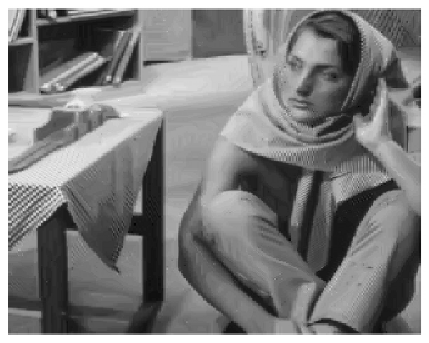}}
\caption{Reconstruction of Barbara from 20\% of measurements in Fourier domain. }
\label{fig:barbara20}
\end{figure*}

\begin{figure*}[h]
\centering
\subfigure[Original image]{\includegraphics[width=8cm]{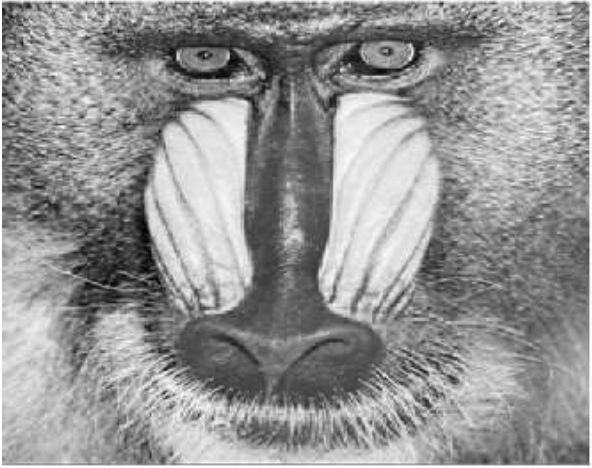}}
\subfigure[Back-projection, $7.35$]{\includegraphics[width=8cm]{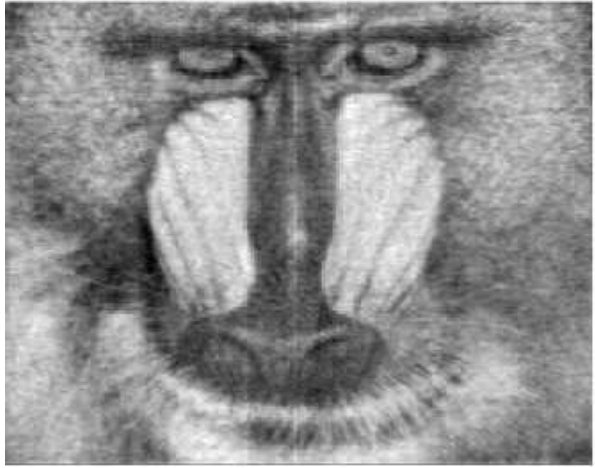}}
\\
\centering
\subfigure[proposed local method, $9.14$ dB]{\includegraphics[width=8cm]{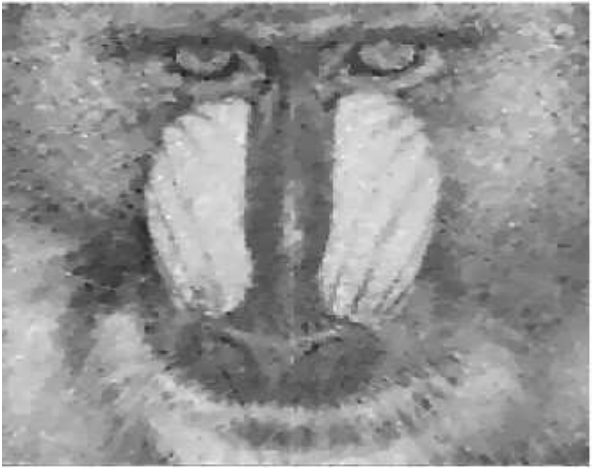}}
\subfigure[proposed non local method, $9.74$ dB]{\includegraphics[width=8cm]{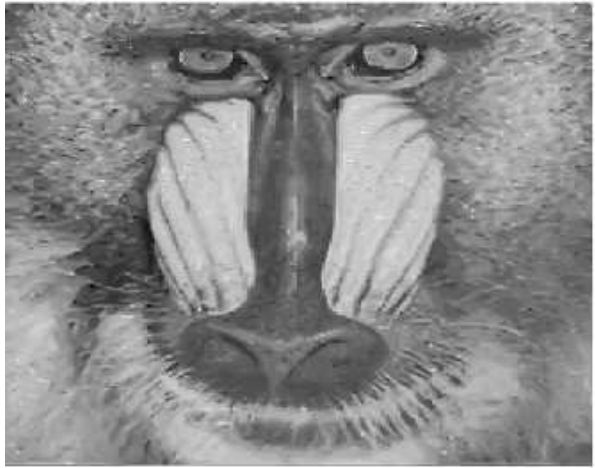}}
\caption{Reconstruction of baboon from 20\% of measurements in Fourier domain. }
\label{fig:baboon20}
\end{figure*}

\begin{figure*}[h]
\centering
\subfigure[Original image]{\includegraphics[width=8cm]{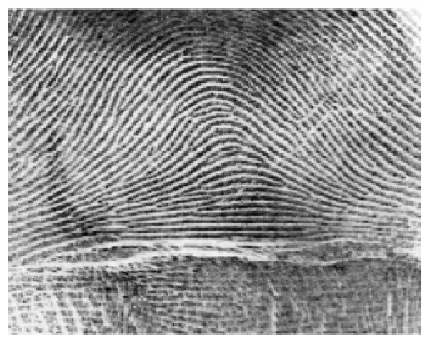}}
\subfigure[Back-projection, $4.69$ dB]{\includegraphics[width=8cm]{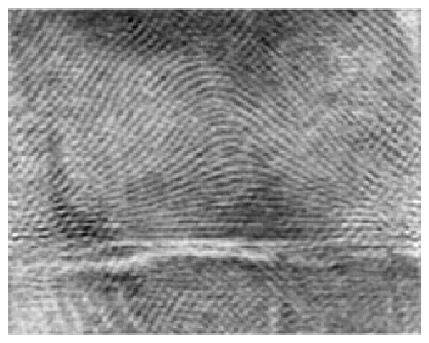}}
\\
\centering
\subfigure[proposed local method, $5.70$ dB]{\includegraphics[width=8cm]{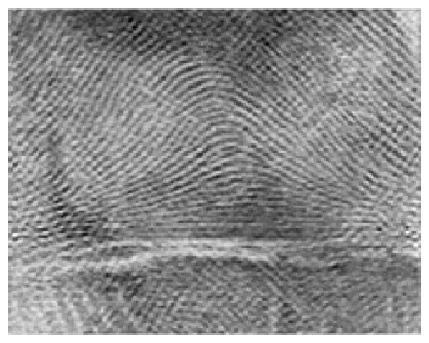}}
\subfigure[proposed non local method, $9.07$ dB]{\includegraphics[width=8cm]{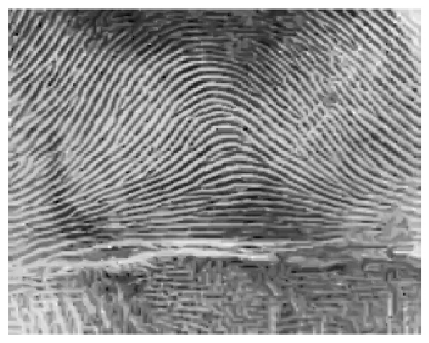}}
\caption{Reconstruction of fingerprint from 20\% of measurements in Fourier domain.}
\label{fig:fingerprint20}
\end{figure*}

In a second set of experiments, we tested our method with natural images containing textures, where edge detection by itself is a difficult task and the images can not be considered piecewise constant. With these images, the local regularization looses all texture information, while the non-local approaches can recover repetitive patterns and better exploit the geometrical information of the image. Results with our method are presented in table \ref{tab:experiments_textured}, with some of the reconstructed images shown in figures \ref{fig:lena12}-\ref{fig:fingerprint20}.
\\
A quantitative comparison of the different methods with textured images is presented in table \ref{tab:experiments_textured}. We observe that the inclusion of an edge detector in edge-guided CS does not improve the TV reconstruction because the partially reconstructed images are not accurate enough to detect edges and the weighted TV term of edge-guided CS encourages edges in wrong positions. That effect is not observed in our method because it is additive and not multiplicative and it exploits the directional information of the regularized normals, which can partially capture texture information better than an edge detector. As a consequence, our local method always outperforms the TV reconstruction and edge-guided CS methods. For the non local regularizations our method outperforms non local TV, but the gain in some cases is negligible (fingerprint and baboon images for a ratio of measurements $\frac{m}{n} = 12\%$ or $20\%$). In fact, the non local methods require a good estimate of the reconstruction to initialize the non local gradient and divergence operators. Since our method requires both gradient and divergence to estimate the non local normals and align them with the reconstruction, we can only improve the non local TV reconstruction when the initialization (in our case we use the standard TV solution) has a minimum level of accuracy.  The fact that more measurements are required for the fingerprint of baboon images is coherent with CS theory, as these images have finer details and are less sparse than Lena or Barbara in terms of total variation. In figure \ref{fig:lena12}-\ref{fig:lena20} we can qualitatively observe the advantages of our method in comparison to local and non-local TV reconstruction for textured images. In the local case we avoid the staircase effect, which is clearly visible in the TV reconstruction of Lena's cheeck. In the non local case, we also capture better slowly varying textures changes, see for instance the different shadows in Lena's skin or hat. In both cases this improvement is due to the regularization of the level set normals of the image, which we exploit for CS recovery with our two step procedure.

\begin{table}[h]
\centering
\tabcolsep 3pt
 \begin{tabular}{|l | c | c c c | c c |}
 \hline
  &  & \multicolumn{3}{c|}{local CS} & \multicolumn{2}{c|}{ non-local CS }\\
$\frac{m}{n}$& image & TV  & edge CS  & normal CS & TV  & normal CS \\
\hline   
&             Lena 			&  14.53 dB &	14.47 dB & \textbf{14.86 dB} & 15.82 dB & \textbf{16.79 dB} \\
12\%&    Barbara 			&  13.35 dB & 	13.31 dB & \textbf{13.59 dB} & 15.00 dB & \textbf{15.52 dB} \\ 
&  fingerprint  &  4.13 dB & 4.11	dB & \textbf{4.13	dB} & 5.97	dB & \textbf{5.98 dB} \\
& 			baboon				&  7.40 dB & 7.25 dB & \textbf{7.40 dB} & 7.65 dB & \textbf{7.65 dB} \\ 
\hline
&             Lena 			& 18.44 dB & 18.36 dB & \textbf{19.27 dB} & 19.95 dB & \textbf{21.09 dB} \\
20\%&    Barbara 			& 16.71 dB & 16.62 dB & \textbf{17.13 dB} & 18.37 dB & \textbf{18.93 dB}\\ 
&  fingerprint  &  5.70 dB & 5.62 dB & \textbf{5.70 dB} & 9.03 dB & \textbf{9.07 dB} \\
& 			baboon				&  9.13 dB & 8.91 dB & \textbf{9.14 dB} & 9.63 dB & \textbf{9.74 dB} \\ 
\hline   
&             Lena 			&  25.39 dB &	25.30 dB & \textbf{26.71 dB} & 26.39 dB & \textbf{27.51 dB} \\
39\%&    Barbara 			&  20.83 dB & 20.68 dB & \textbf{21.36	dB} &  24.68 dB & \textbf{25.33 dB} \\ 
&  fingerprint  &  12.02 dB & 11.84 dB & \textbf{12.03 dB} & 14.52 dB & \textbf{14.56 dB} \\
& 			baboon				&  13.30 dB & 13.14 dB & \textbf{13.41 dB} & 13.44 dB & \textbf{13.82 dB}\\ 
\hline
\end{tabular} 
\caption{Comparison of CS reconstruction for textured images. The first three columns show the results with the standard TV term in the regularization: TV stands for the model \eqref{eq:CSmin}, edge CS for \eqref{eq:edgeCS} and normal CS for our method \eqref{eq:CSnormal}. The last two columns correspond to the definition of NL TV: NL-TV corresponds to the standard non local CS recovery \eqref{eq:NLCSmin} and NL normal CS for the proposed non local method \eqref{eq:NLCSnormal}. }
\label{tab:experiments_textured}
\end{table}

\section{Conclusions} \label{sec:conclusions}
We propose a normal guided compressed sensing recovery method to recover images of higher qualities from fewer measurements. The normal vectors of image level curves have been exploited in denoising and inpainting algorithms, but in compressed sensing this information is embedded in the measurements and state-of-the-art recovery algorithms have just neglected it. To extract this geometric information we alternatively estimate the normals of the image level set curves and then improve the compressed sensing reconstruction matching the estimated normals, the compressed sensing measurements and the sparsity constraints. With our approach, the geometric information of level contours are incorporated in the image recovery process. The proposed method is also extended to non local operators to recover textured images and could also be applied to improve existing non local denoising and deblurring methods. Our numerical experiments show that the proposed method improves image recovery in several ways: it is able to recover sharp edges as well as smoothly varying image regions, avoiding the staircase effect in the case of total variation reegularization; it is robust to noise and the sparsity of the signal and relies on  efficient minimization techniques to obtain a fast and  easy-to-code algorithm.

\bibliographystyle{IEEEtran}
\bibliography{bibfile}
\end{document}